\documentclass[runningheads]{llncs}
\usepackage{graphicx}
\usepackage{bbm}
\usepackage{tikz}
\usepackage{comment}
\usepackage{amsmath,amssymb} 
\usepackage{color}
\usepackage{xfrac}
\usepackage{algorithm}
\usepackage{algpseudocode}
\usepackage[accsupp]{axessibility}  
\usepackage{blindtext}
\usepackage{wrapfig}
\usepackage{cite}
\usepackage{microtype}
\usepackage{tabularx}
\usepackage{booktabs}
\usepackage{multirow}

\usepackage{setspace}
\usepackage{enumitem}
\usepackage{multirow}
\usepackage{wrapfig}
\usepackage{colortbl}
\usepackage[normalem]{ulem}

\let\llncssubparagraph\subparagraph
\let\subparagraph\paragraph
\usepackage[compact]{titlesec}
\let\subparagraph\llncssubparagraph
\titlespacing{\section}{0pt}{3ex plus 1ex minus .2ex}{2ex minus .2ex}
\titlespacing{\subsection}{0pt}{2ex plus 1ex minus .2ex}{1ex minus .2ex}

\setlist[itemize]{align=parleft,left=0pt}

\definecolor{azure(colorwheel)}{rgb}{0.0, 0.5, 1.0}
\definecolor{nicegreen}{rgb}{0.0, 0.7, 0.1}
\definecolor{ashblue}{rgb}{0.36, 0.54, 0.66}
\definecolor{ashgrey}{rgb}{0.7, 0.75, 0.71}
\definecolor{applegreen}{rgb}{0.55, 0.71, 0.0}
\definecolor{yw}{rgb}{0.784, 0.003, 0.313}
\definecolor{ywg}{rgb}{0.9960, 0.8984, 0.5859}
\definecolor{jy}{rgb}{0.58, 0, 0.827}
\definecolor{cornellred}{rgb}{0.7, 0.11, 0.11}
\definecolor{darkcyan}{rgb}{0.0, 0.55, 0.55}
\definecolor{CuGray}{gray}{0.9}
\definecolor{airforceblue}{rgb}{0.36, 0.54, 0.66}
\definecolor{rev}{rgb}{0.784, 0.003, 0.313}
\definecolor{pink}{cmyk}{0, 0.7808, 0.4429, 0.1412}
\definecolor{amethyst}{rgb}{0.6, 0.4, 0.8}
\definecolor{black}{rgb}{0.0, 0.0, 0.0}
\definecolor{tb3_yellow}{rgb}{0.996, 1.0, 0.6}
\definecolor{tb3_orange}{rgb}{0.980, 0.8, 0.604}
\definecolor{tb3_red}{rgb}{0.972, 0.6, 0.6}
\definecolor{dimgray}{rgb}{0.41, 0.41, 0.41}
\definecolor{brickred}{rgb}{0.8, 0.25, 0.33}
\definecolor{bleudefrance}{rgb}{0.19, 0.55, 0.91}
\definecolor{blue(ncs)}{rgb}{0.265, 0.445, 0.765}

\newcolumntype{g}{>{\columncolor{CuGray}}c}
\newcolumntype{z}{>{\columncolor{CuGray}}l}

\renewcommand{\paragraph}[1]{\vspace{1mm}\noindent\textbf{#1.}\,\,}
\newcommand{\colorref}[1]{{\color{cornellred}{#1}}}

\newcommand{\grc}[1]{\textcolor{dimgray}{#1}}

\newcommand{\clipscore}[1]{\textcolor{blue(ncs)}{#1}}

\usepackage{xspace}

\newcommand*\samethanks[1][\value{footnote}]{\footnotemark[#1]}
\makeatletter
\def\@fnsymbol#1{\ensuremath{\ifcase#1\or *\or \dagger\or \ddagger\or
   \mathsection\or \mathparagraph\or \|\or **\or \dagger\dagger
   \or \ddagger\ddagger \else\@ctrerr\fi}}
\makeatother

\def\onedot{.\@\xspace}
\def\eg{\emph{e.g}\onedot} 
\def\ie{\emph{i.e}\onedot}

\def\etal{\emph{et al}\onedot}

\newcommand{\Sref}[1]{Sec.~\ref{#1}}
\newcommand{\Eref}[1]{Eq.~(\ref{#1})}
\newcommand{\Fref}[1]{Fig.~\ref{#1}}
\newcommand{\Tref}[1]{Table~\ref{#1}}

\newcommand{\be}{\begin{eqnarray}}
\newcommand{\ee}{\end{eqnarray}}
\newcommand{\bee}{\begin{eqnarray*}}
\newcommand{\eee}{\end{eqnarray*}}

\newcommand{\matrixb}{\left[ \begin{array}}
\newcommand{\matrixe}{\end{array} \right]}

\newcommand{\argmax}{\operatornamewithlimits{\arg \max}}

\newcommand{\para}[1]{\paragraph{#1}}
\usepackage[pagebackref,breaklinks,colorlinks=True,citecolor=cornellred,linkcolor=cornellred,bookmarks=false]{hyperref}

\begin{document}

\pagestyle{headings}
\mainmatter
\def\ECCVSubNumber{3229}

\title{CLIP-Actor: Text-Driven 
Recommendation \\ and Stylization 
for Animating Human Meshes}

\titlerunning{CLIP-Actor}
\authorrunning{Youwang et al.}
\author{
Kim Youwang\inst{1}\thanks{Authors contributed equally to this work.}
\qquad
Kim Ji-Yeon\inst{2}\samethanks
\qquad
Tae-Hyun Oh\inst{1,2}\thanks{Joint affiliated with Yonsei University, Korea.}
}

\institute{
Department of ${}^1$EE \& ${}^2$CiTE, POSTECH, Korea\\
\email{\{youwang.kim, jiyeon.kim, taehyun\}@postech.ac.kr}\\
\url{https://clip-actor.github.io}
}

\maketitle

\begin{abstract}
We propose CLIP-Actor, a
text-driven motion recommendation and neural mesh stylization system for human mesh animation.
CLIP-Actor animates a 3D human mesh to conform to a text prompt by recommending a motion sequence and 
optimizing mesh style attributes.
%
We build a text-driven human motion recommendation system by leveraging 
a large-scale
human motion dataset with language labels.
Given a natural language prompt, CLIP-Actor 
suggests a text-conforming human motion
in a coarse-to-fine manner.
Then, our novel zero-shot neural style optimization
detailizes and texturizes 
the recommended 
mesh sequence 
to conform to the prompt in a temporally-consistent and pose-agnostic manner.
This is distinctive in that prior work fails to generate plausible results when the
pose of an artist-designed mesh 
does not conform to the text 
from the beginning.  
%
We further propose the spatio-temporal view augmentation and mask-weighted embedding attention, 
which stabilize the optimization process by leveraging multi-frame human motion and 
rejecting poorly rendered views.
We demonstrate that 
CLIP-Actor produces plausible and 
human-recognizable 
style 3D human mesh in motion with 
detailed geometry and texture solely from a natural language prompt.


\end{abstract}

\section{Introduction}
Manual generation of animatable and detailed 3D avatars is cumbersome and requires time-consuming efforts with intensive labor and pain of creation.
To reduce such burdens, many attempts have been introduced to automate such processes 
\cite{Gafni_2021_CVPR,Canfes2022TextAI,Saito:CVPR:2021,inversesim, Li2021NClothP3,Bhatnagar2019MultiGarmentNL,bozic2020neuraltracking,bozic2021neuraldeformationgraphs,palafox2021npm}. 
%
%
Furthermore, 
highly deformable human bodies make it more 
challenging to design temporally-consistent detailed geometries and textures.
%
This 
process may be fully automated by 
%
%
text-guided 3D avatar generation, \ie,
making a machine understand
the human text prompt to create a 3D avatar amenable to the prompt.
%
%
Text-guided 3D avatar generation can be widely applied to machine-created media, such as virtual human animation~\cite{Saito:CVPR:2021}, 
language-driven robot task planning~\cite{yoonICRA19, shree2021exploiting}, and movie script visualization~\cite{Hanser2009SceneMakerIM}.

Our key intuition 
is 
from the text-visual coupled understanding of humans.
For example, when an actor reads a script for a play, 
the actor brings up an image of
gestures, tone of speech, and clothes to her/his mind
following the context described in the script.
We believe such text-visual coupled imagination can be a breakthrough for 
accelerating machine-created media, \eg, stylized 3D humans in motion.
We can embody it to the machine 
by leveraging the text-image joint embedding space of CLIP~\cite{radford2021learning}.
%
With the representational power of the CLIP embedding space, 
the similarity measure between text and image
provides concrete signals in building text-to-3D human meshes in motion.

\begin{figure}[t]
\centering
  \includegraphics[width=0.9\linewidth]{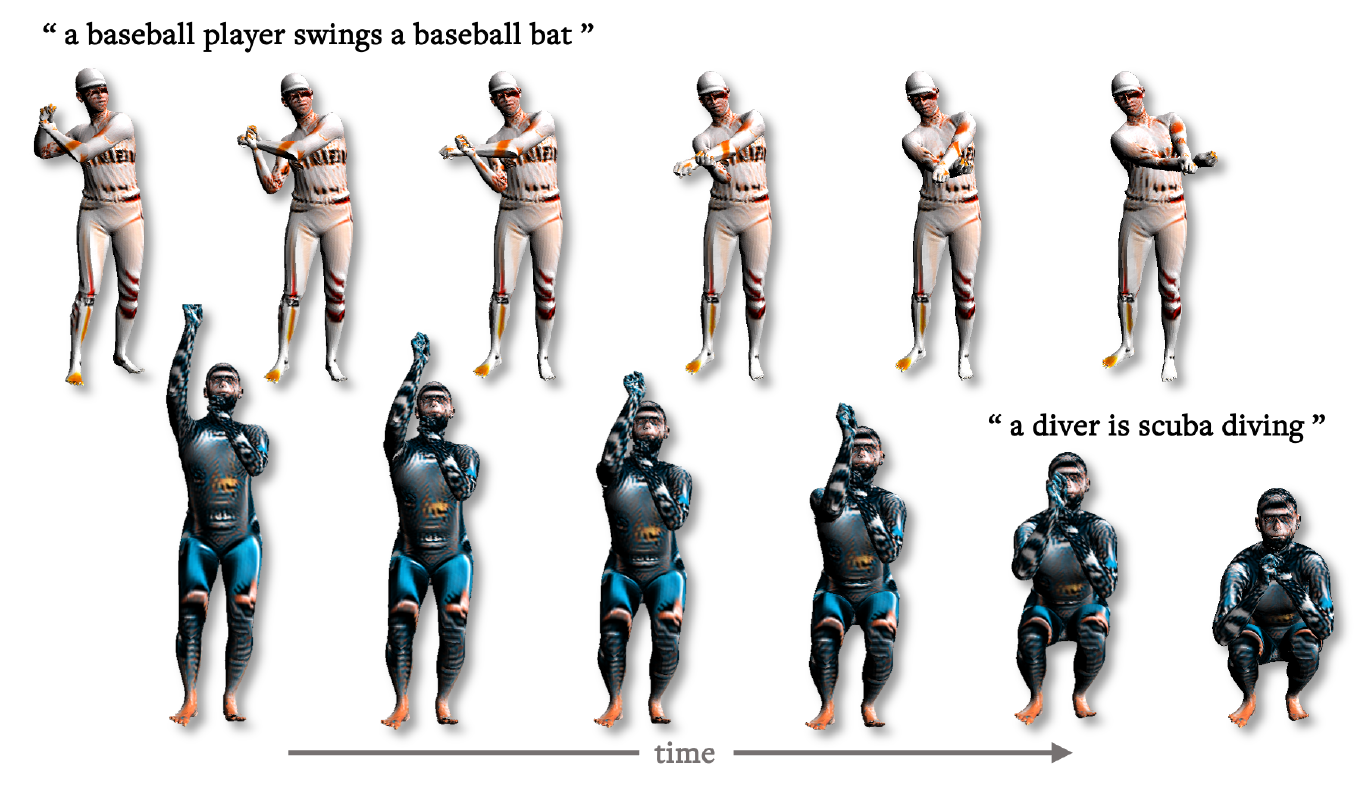}
   \caption{{\textbf{CLIP-Actor.} Given an input text prompt, CLIP-Actor recommends the 
   best matching human meshes in motion
   and iteratively stylizes them by synthesis-through-optimization.
   CLIP-Actor can detail and texturize not just a single mesh frame
   but a short action clip by optimizing temporal-consistent and pose-agnostic style attributes.
   }}
\label{fig:demo}
\end{figure}

In this work,
we propose \textbf{CLIP-Actor}, an automated framework of text-driven recommendation and stylization of animating 3D human meshes. 
Given a text prompt describing human action and style, 
CLIP-Actor crafts a short clip of animated human meshes conforming to the prompt (see \Fref{fig:demo}).
%
Our method is free from extra artist-designed 3D mesh inputs since 
it searches meshes in motion from a database
that strongly correlates with the given query text. 
%
The CLIP-Actor then detailizes and texturizes the mesh sequence by optimizing our proposed Decoupled Neural Style Fields (DNSF)
in a pose-agnostic manner.
The objective of the optimization is to maximize 
the correlation between the input text prompt and 2D rendered images of the stylized 3D mesh.
We optimize DNSF with spatio-temporally augmented rendered images 
and provide an initial content mesh with a multi-modal sampling strategy. 
%
Moreover, we propose 
mask-weighted embedding attention 
for stable neural optimization.
%
We demonstrate that CLIP-Actor can stylize visually and physically plausible 
3D human meshes in motion with various text descriptions in zero-shot.
%

\noindent We summarize our main contributions as follows:

\begin{itemize}
  \item[$\bullet$] 
  We propose CLIP-Actor, a text-driven animated human mesh synthesis system.
  \item[$\bullet$]
  We propose a hierarchical text-driven human motion recommendation module
  that utilizes 
  fine-grained textual semantic matching to capture 
  visual and textual cues within the text prompt. 
  \item[$\bullet$]
  Our novel decoupled neural style field (DNSF) learns style attributes of 
  the human meshes in motion in a temporally consistent and pose-agnostic manner.
  
  \item[$\bullet$]
  We further develop novel methods to improve the convergence of 
  the text-driven neural DNSF optimization:
  multi-modal content mesh sampling, spatio-temporal view augmentation, and mask-weighted 
  embedding attention. 
\end{itemize}

\section{Related Work}
Our work is closely related to text-driven 3D object contents and style manipulation. 
Multi-modal object stylization has been mainly studied using learned multi-modal embedding space, 
such as CLIP~\cite{radford2021learning} and 3D content/style manipulation methods.
We briefly review these lines of research.

\para{Text-driven visual data
manipulation}
Recent advances in learned text and image joint embedding space~\cite{du2021gem,radford2021learning} have
lit a fire in
research about the style manipulation of images and 3D objects.
CLIP embedding space is learned with abundant natural images and texts and was originally developed for zero-shot image and language analysis tasks.
Interestingly, its representation turns out powerful enough to manipulate visual data with intuitive text guides.
For images, text conditional image generation \cite{Frans2021CLIPDrawET,Nichol2021GLIDETP,Kim2021DiffusionCLIPTD,Kwon2021CLIPstylerIS,Gal2021StyleGANNADACD} has been notably advanced by CLIP. 
%
A representative work, StyleCLIP~\cite{Patashnik_2021_ICCV}, manipulates an input image by 
optimizing over its latent code of a pre-trained generative model given a natural language text-prompt. 
CLIPDraw~\cite{Frans2021CLIPDrawET} synthesizes images with text guidance by optimizing 
the parameters of a set of curves via gradient descent.

Analogous to the image domain, several works~\cite{jain2021dreamfields,text2mesh} extend the manipulation target domain to 
3D objects by leveraging the advances in differentiable rendering~\cite{Niemeyer2020CVPR,mildenhall2020nerf,kato2020differentiable,ravi2020pytorch3d,barron2021mipnerf360}.
The differentiable rendering technique enables seamless gradient flow from 2D rendered images to their 3D objects, 
allowing CLIP to bridge between language and 3D modalities through 2D images.
Dream Fields~\cite{jain2021dreamfields} generates a 3D structure using implicit representation in free space, given a text prompt. 
It exploits no structural prior knowledge to learn or manipulate 3D contents.
This allows flexible content exploration with novel styles but often results in abstract visual contents.
As another concurrent work, Michel~\etal\cite{text2mesh} propose Text2Mesh, a CLIP-guided optimization method to manipulate the given fixed source mesh styles to conform to the target text condition prompt.
In contrast to Dream Fields, since Text2Mesh stylizes a 3D object over the displacement and its texture map defined on a fixed,
T-posed template human mesh, it imposes strong structural prior.
It demonstrates plausible and interesting styles and textures of meshes given a text prompt. 
However, we observed that when the given template mesh is hard to conform to the given text prompt, it produces undesirable stylization;
\eg, the text containing detailed human action produces a failure of stylization when the pose of the given human template mesh and the action are not conformed to each other.

We focus on 
animating human meshes with details and styles according to the input text prompt.
We exploit the parametric human mesh model to disentangle style from geometric contents, \ie, pose.
Such disentanglement enables the pose, detail, and style of human mesh to conform to the input text sequentially.
This enables to stably manipulate 3D human objects, better conforming to the input text prompt from action to style.

\para{Text-driven human motion manipulation}
Many recent approaches have been studied to generate human body motion with given natural language descriptions.
%
One line of work~\cite{ahn:Text2Action:icra18,Ahuja2019Language2PoseNL,Plappert2018LearningAB,Ghosh2021SynthesisOC,lin:vigil18} guides the machine to translate natural language descriptions in a sequential manner and generate human skeletal motions using recurrent neural models.
%
Another line of
work~\cite{petrovich21actor,guo2022action2video,chuan2020action2motion} 
generates human motion conditioned on the limited number of closed-set action categories.
%
%
CLIP-Actor focuses on textual and visual semantics in a whole sentence and can tackle 
various natural language descriptions.

Recently, MotionCLIP~\cite{tevet2022motionclip} and TEMOS~\cite{petrovich22temos} propose to learn the natural language conditioned mesh motion generation. 
MotionCLIP learns the human motion autoencoder and makes its latent space compatible with CLIP text and image space using semantic similarity. 
%
%
Similarly, TEMOS learns generative human mesh motion latent space with 
transformer-VAE~\cite{petrovich21actor,trans_vae,recurrent_katerina} and aligns it with natural language latent space via DistilBERT~\cite{Sanh2019DistilBERTAD}, thus composing the cross-modal motion latent space.
%
%
While both methods focus on the latent space to capture textual and visual semantics of natural language descriptions, CLIP-Actor directly maps the descriptions to realistic motion using a recommendation system. 
Moreover, our detailed volumetric meshes are stylized with appearance attributes 
much more expressive than those of the aforementioned methods.
%
%

\para{Texture and geometric stylization of human mesh in motion}
Aside from
3D mesh pose,
recent work has added different levels of details
to bare human meshes, \eg, cloth modeling or texture color. 
The separate modeling of human and cloth meshes~\cite{inversesim, Li2021NClothP3,Bhatnagar2019MultiGarmentNL}, 
the neural extension of the parametric human mesh model~\cite{Burov2021DynamicSF, bhatnagar2020ipnet,ma2020cape},
the neural parametric approach~\cite{bozic2020neuraltracking,bozic2021neuraldeformationgraphs,palafox2021npm},
and the neural implicit approach~\cite{saito2019pifu,saito2020pifuhd,Wang2021NEURIPS} 
show promising clothed human mesh 
results from the given human scans, but without surface colors.
Those works deal with texture and geometric styles separately. 
Recently, Saito~\etal\cite{Saito:CVPR:2021} propose a weakly-supervised way to recover both texture and geometric styles.

None of these methods can generate diverse color and cloth details of human motion in a zero-shot manner, \eg, with only a text guide. 
We present a novel text-driven recommendation, detailization, and texturization of animating human meshes in zero-shot, where human meshes in motion with 
texture and geometric details are generated from the machine's imagination without the task dataset.

\section{CLIP-Actor: An Overview}
\label{sec:3_ProblemStatement}
\begin{figure*}[t]
\centering
   \includegraphics[width=\linewidth]{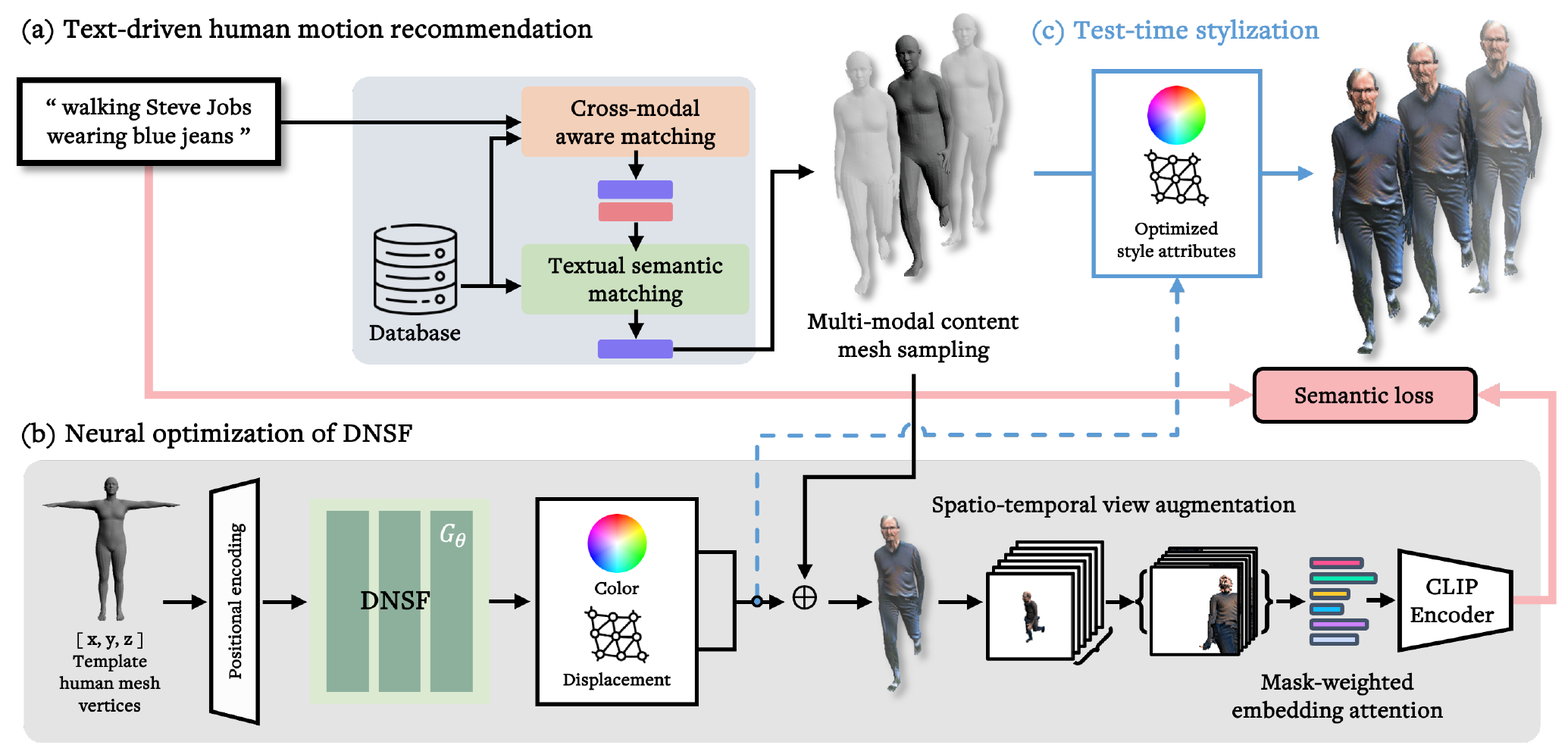}
   \caption{{\textbf{Overall architecture}. Given a text description of the human action, 
   the text-driven human motion recommendation module finds the best semantically matched motion sequence from the motion database~\cite{BABEL:CVPR:2021}. 
   Content meshes are then sampled in a multi-modal context. 
   Decoupled Neural Style Field takes a T-posed human mesh and learns text-driven style attributes,
   which are then applied to the content meshes.  
   We apply spatio-temporal view augmentation
   and weight rendered images to guide the neural optimization with similarity among rendered images and the text.}}
\label{fig:network} 
\end{figure*}
Our goal is to visualize 3D motion that conforms to the input description by stylizing mesh with the color and displacement of its vertices. 
For example, consider a natural language prompt, ``walking Steve Jobs wearing blue jeans."
Instead of preparing extra fixed 3D mesh inputs, 
our model obtains a sequence of 3D meshes that conforms to the
input prompt, \ie, walking, by retrieving a motion sequence from a dataset, \eg, BABEL~\cite{BABEL:CVPR:2021}.
The retrieved mesh sequence becomes the \emph{``content"} of our mesh stylization. 
We then grant the characteristics, \eg, cloth, hair, to the meshes by optimizing the neural model to learn the color and displacement of the mesh vertices.
%
Finally, our model generates a short clip of walking Steve jobs wearing blue jeans (see \Fref{fig:network}).

Formally, given a text prompt ${y}$, we retrieve a sequence of pose parameters
$\mathbf{R}_{1:T} = [\mathbf{R}_1,\dots,\mathbf{R}_{T}]$ of SMPL~\cite{SMPL:2015,MANO:SIGGRAPHASIA:2017,SMPL-X:2019} for duration $T$. 
In a single frame $t$, mesh vertices $\mathbf{M}_t$ can be acquired with a linear mapping as:
$\mathbf{M}_t = \mathcal{M}(\mathbf{R}_t, \boldsymbol{\beta}_t), \,\, \forall{t}\in\{1,{\dots},T\},$
where $\mathbf{R}_t$ denotes the pose parameters, and $\boldsymbol{\beta}_t$ 
the shape parameters for a human mesh. 
Then, a single mesh at frame $t$ is represented by the faces $F$,
and the 3D mesh vertices
{$\mathbf{M}_t\in{\mathbb R}^{V\times3}$},
where $V$ is the number of vertices. 
Since SMPL mesh faces $F$ for every frame are identical with given triangulation, 
we represent a single mesh using the mesh vertices, $\mathbf{M}_t$.
Hence,
$\mathbf{M}_{1:T}=[\mathbf{M}_1,\dots,\mathbf{M}_{T}]$ denotes a full sequence of human meshes and is taken to our decoupled neural style field (DNSF) as \emph{``content."}
The DNSF 
then learns \emph{``style,"} 
\ie, color and displacement, of mesh vertices and produces a sequence of textured mesh $\mathbf{M}^{*}_{1:T}$.

\section{Text-driven Human Motion Recommendation}
\begin{figure*}[t]
\centering
   \includegraphics[width=\linewidth]{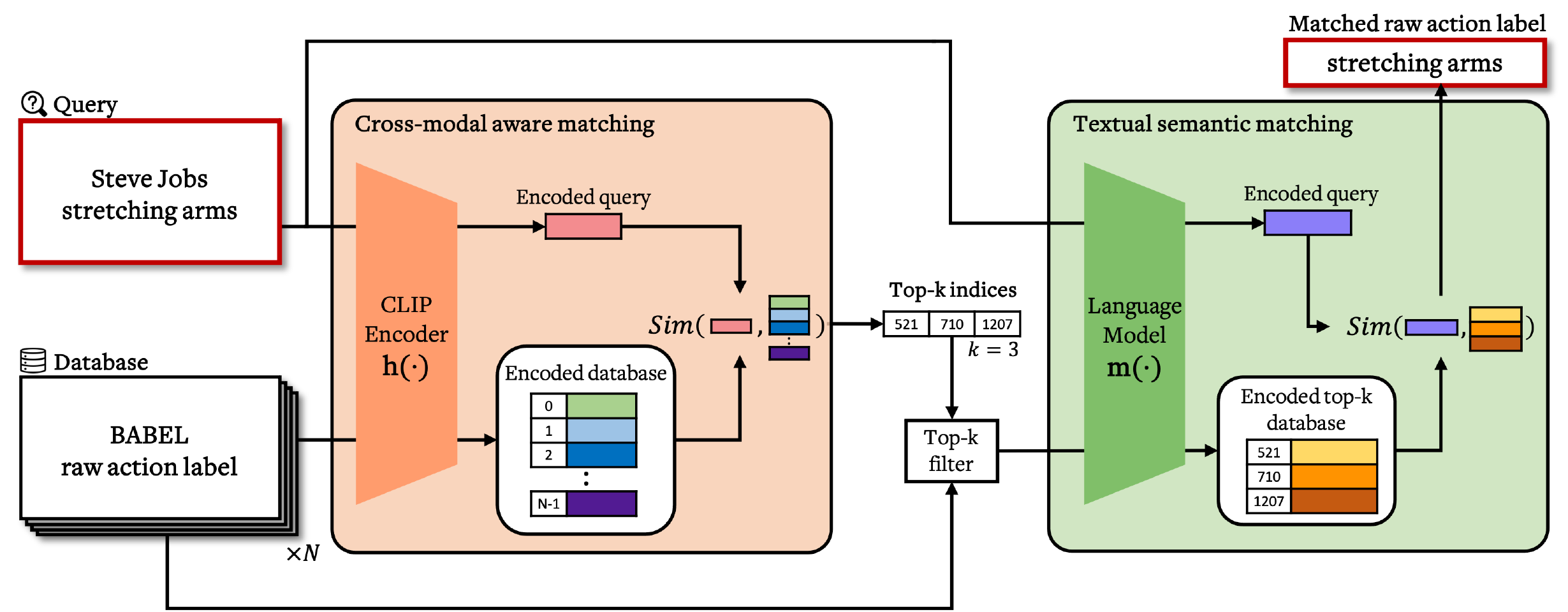}
   \caption{{\textbf{Hierarchical multi-modal motion retrieval}. Given text prompt as a query, our retrieval module finds the most relevant raw action label from the database~\cite{BABEL:CVPR:2021}. First, the query and all of the raw action labels are encoded by CLIP text encoder $\textbf{h}(\cdot)$, and we measure similarity between them.
   Top-$\mathtt{k}$ indices of raw action labels are selected, 
   and corresponding raw action labels are retrieved by the $\mathtt{top{-}k}$ filter. 
   The language model encoder $\textbf{m}(\cdot)$ vectorizes the query and top-$\mathtt{k}$ action labels.
   The highest-scored raw action label is retrieved as the final matched result for the input text prompt.}}
\label{fig:retrieval}
\end{figure*}
\label{sec:4_motion_recommendation}
In this section, we propose a motion recommendation module to obtain a motion sequence that conforms to the text prompt.
We recommend the motion by retrieving visually and textually relevant action labels from the dataset.

\para{Hierarchical multi-modal motion retrieval}
We propose a hierarchical multi-modal motion retrieval module
to obtain a motion sequence corresponding to the given prompt (see \Fref{fig:retrieval}). 
We utilize a large scale 
human motion dataset with language labels, BABEL~\cite{BABEL:CVPR:2021}, containing frame-level aligned SMPL pose parameters and 
raw action labels.
Given the text prompt as a query, 
the raw action label that is visually and linguistically associated with the query 
is matched through our motion recommendation system.
We design a two-stage retrieval; cross-modal aware matching and textual semantic matching.
The hierarchical matching enables CLIP-Actor to catch visual-language (cross-modal) aware contexts and linguistic semantics. 
These comprehensive matching modules hand over good initial content to the subsequent neural mesh stylization.

\para{Cross-modal aware matching}
Cross-modal aware matching 
finds the action labels similar to the input text prompt on the joint image-text space.
%
We prepare the database, \ie, a set of raw action labels $\mathcal{A}$, gathered from BABEL~\cite{BABEL:CVPR:2021}.
%
We retrieve a set $\mathcal{A}_\mathtt{k}{\scriptstyle \subset} \mathcal{A}$ of the raw action label $a_i\in \mathcal{A}$, given a
text prompt 
$y$ as:
%
\begin{equation}
\label{eq:clip_sim}
    \mathcal{A}_\mathtt{k} = \mathtt{top{-}k}[\mathcal{S}(\mathbf{h}(a_i), \mathbf{h}(y))], 
    \quad \textrm{where\,\,\,} \mathcal{S}(\mathbf{x}, \mathbf{y}) = \dfrac
    {\mathbf{x}^{\top} \mathbf{y}}
    {
    \lVert \mathbf{x}\rVert_{\scriptscriptstyle 2}
    \lVert \mathbf{y}\rVert_{\scriptscriptstyle 2}
    },
\end{equation}
$\mathbf{h}(\cdot)$ is the pre-trained CLIP text encoder, 
and $\mathtt{top{-}k}[\cdot]$ denotes a function that returns $\mathtt{k}$ best matches. 
The similarity is measured by the cosine similarity.

Specifically, consider an input prompt
``a man walking backwards."
The cross-modal aware matching vectorizes the input prompt and the action labels using the CLIP text encoder and computes the similarity between them.
The set of matched action labels $\mathcal{A}_\mathtt{k}$ is determined as \{``walking in place," ``walking backward," ``walking laterally"\} and the top-1 matched label is ``walking in place."
Since the CLIP text encoder is learned to focus on words that appear visually, it catches visual semantics (\ie, walking), and all the elements in $\mathcal{A}_\mathtt{k}$ are closely related to the input prompt in visual space. 
However, the text encoder of CLIP is trained with still images instead of videos so that it cannot distinguish fine-grained action (\ie, ``walking in place" vs. ``walking backwards"; because both appear to be the same in a still image).
Thus, we propose textual semantic matching as the following step to compensate for the single-stage retrieval.

\para{Textual semantic matching}
The textual semantic matching finds the most relevant action label with the input prompt by capturing textual semantics in the sentence.
%
We utilize the language expert,
MPNet~\cite{song2020mpnet},
so that our two-stage module can distinguish textual semantics and grammatical structures.
%
The best matching label $a_*$ is retrieved as:
\begin{equation}
\label{eq:sent2vec_sim}
    {a_*} = \argmax\nolimits_{a_j\in \mathcal{A}_\mathtt{k}} \, \mathcal{S}(\mathbf{m}(a_j), \mathbf{m}(y)), \quad 
\end{equation}
where $\mathbf{m}(\cdot)$ denotes the pre-trained MPNet encoder.
Again, consider our 
above example of ``a man walking backwards."
The top-$\mathtt{k}$ action labels are re-ranked, and the most similar action label ``walking backward" is retrieved as a final result.
The sequence of meshes $\mathbf{M}_{1:T}$ associated with the retrieved action label is passed to the following 
neural mesh stylization pipeline as the content mesh sequence.

\section{Decoupled Stylization of Human Meshes in Motion}

\label{sec:5_DNSF}
{
We represent a stylized human mesh with
the content mesh 
and the style attributes.
%
In practice, we 
denote the content mesh as $\mathbf{M}_{i}\in\mathbb{R}^{V \times 3}$
sampled from the retrieved human motion sequence of T frames, $\mathbf{M}_{1:T}$.
%
The mesh's surface style attributes 
$(\mathbf{c}, \mathbf{d})\in \mathbb{R}^{V\times3}\times \mathbb{R}^{V}$ 
are interpreted as 
the per-vertex RGB color 
and per-vertex displacement,
which are applied over surfaces via given triangulation.
%
The Neural Style Field~\cite{text2mesh} 
takes a fixed static mesh as input and learns the
style attributes with a multi-layer perceptron (MLP).
Michel \etal\cite{text2mesh} claim that this implicit formulation tightly couples the style field to the source mesh.
However, since Neural Style Field
takes a single posed mesh at a time, a significant number of MLPs are required to stylize a sequence of human meshes.
}


\para{Decoupled Neural Style Field}
Instead, we introduce \emph{Decoupled Neural Style Field} (DNSF).
We propose to rather decouple the style field from the content mesh so that we need only one neural network to learn style attributes for the meshes in motion.
Specifically, we first map the style attributes from the 
template human mesh $\mathbf{M}_{c}$
and merge it with the content meshes $\mathbf{M}_{i}, \forall{i}\in\{1,\dots,T\}$, right before rendering (refer to \Fref{fig:network}).
DNSF can achieve the same mesh stylization as the basic Neural Style Field
while effectively decoupling the style from the content mesh.
In practice,
parameterized as an MLP $G_{\theta}$, DNSF maps the vertices 
on the template human mesh, \ie, T-posed SMPL $\mathbf{M}_{c}$, 
to style attributes $\mathbf{c}$ and $\mathbf{d}$, in a pose-agnostic manner as:
\begin{equation}
\textrm{DNSF:}\quad
G_{\theta}(\mathbf{M}_{c}) \mapsto \{ \mathbf{c}, \mathbf{d}\}.
\end{equation}

We also employ the Fourier feature-based positional encoding to the mesh vertices,
which helps the style field to cover higher frequency details~\cite{tancik2020fourfeat}. 
In detail,
the MLP $G_{\theta}$ gets the positional encoded feature as input
and outputs 
per-vertex RGB value, 
$\mathbf{c}\in[0,1]^{V\times3}$, and 
per-vertex displacement value, 
$\mathbf{d} \in [-0.1,0.1]^{V}$, along the per-vertex normal direction. 
The predicted style attributes are then applied
to the content posed mesh $\mathbf{M}_{i}$
to produce the stylized human mesh $\mathbf{M}^{*}_{i}$.

\para{Text-driven DNSF optimization}
The core of the text-driven DNSF optimization is 
to maximize the semantic correlation between the visual mesh observation and the input text prompt. 
However, we cannot directly utilize CLIP to 
measure the semantic correlation with the created 3D mesh itself because the CLIP visual encoder is designed and trained only for 2D images.

We leverage an intuitive idea that
the observations of a 3D object can be described similarly from any viewpoint~\cite{Jain_2021_ICCV,jain2021dreamfields,wang2021clip}.
%
To utilize the representation power of CLIP as a supervision signal, 
we first 
render images of the 3D meshes for input compatibility. 
With randomly sampled $N$ camera poses, $\mathbf{p}=[\mathbf{p}_1,\dots,\mathbf{p}_{N}]$, 
we differentiably render the stylized mesh $\mathbf{M}^{*}_i$ to get
$N$-view 
rendered images $\mathbf{I}^{*}_{ij}$, $\forall j \in \{1,2,\dots,N\}$~\cite{Kaolin2019}.
Thereby,
our main optimization objective, \emph{semantic loss}, is defined with the pre-trained CLIP image and text encoders, $\mathbf{g}(\cdot)$ and $\mathbf{h}(\cdot)$, as:
{
\begin{equation}
    \mathcal{L}_{s}= 
    1 - 
    \dfrac
    {
        \mathbf{\bar{g}}(\mathbf{I}_{i}^{*})^{\top}\mathbf{h}(y)
    }
    {
        \lVert \mathbf{\bar{g}}(\mathbf{I}_{i}^{*})\rVert_{\scriptscriptstyle 2}
        \lVert \mathbf{h}(y)\rVert_{\scriptscriptstyle 2}
    },
    \quad 
    \mathbf{\bar{g}}(\mathbf{I}^{*}_{i}) = 
    \dfrac{1}{N}\sum\nolimits_{j=1}^{N}\mathbf{g}(\mathbf{I}^{*}_{ij}),
    \label{eq:semantic_loss}
\end{equation}
where $y$ denotes the input text prompt, and $\mathbf{g}(\mathbf{I}^{*}_{i})\in\mathbb{R}^{512}$ and $\mathbf{h}(y)\in\mathbb{R}^{512}$ 
the unnormalized CLIP embedding vectors for the image and the text prompt, respectively.
The semantic loss is basically a cosine similarity between the 
normalized mean 
embedding vectors for $N$ rendered images of the stylized mesh $\mathbf{M}^{*}_i$
and the normalized embedding for the input text prompt $y$.
}


\para{Spatio-temporal view augmentation}
{
Prior works show that \emph{spatial} augmentations, 
such as 3D viewpoint or 2D image augmentations,
improve the quality of content generation 
\cite{Frans2021CLIPDrawET,Jain_2021_ICCV,jain2021dreamfields,text2mesh}.
We extend it to \emph{spatio-temporal view augmentation}, where we propose to leverage both multi-view property and human motion originating from the temporal movement.
This naturally diversifies views in a combinatorial way by the spatio-temporal context of human motion. 
%
%
}

{
The strength of DNSF is amplified with the spatio-temporal view augmentation.
Recall that DNSF 
$G_{\theta}$ takes a template SMPL mesh as an input, 
which is pose-agnostic.
%
Therefore, the semantic loss $\mathcal{L}_s$ can be measured with any 
content mesh $\mathbf{M}_{i\in \{1,\dots,T\}}$
in the motion sequence for learning DNSF.
%
One can sample and use the center frame or
the frame that conforms best with the text prompt.
This increases the chance to measure the loss with a view favorable to DNSF learning.
Considering that the na\"ive selection of the content mesh fails to 
generate plausible color and geometric details when it does not conform with the text prompt, 
the content mesh sampling strategy is a crucial design choice. 
}


\para{Multi-modal content mesh sampling} 
One na\"ive way to choose the content mesh to stylize with 
is to randomly select a single mesh within a mesh sequence.
\begin{wrapfigure}[12]{r}{0.40\linewidth}
\centering
    \includegraphics[width=\linewidth]{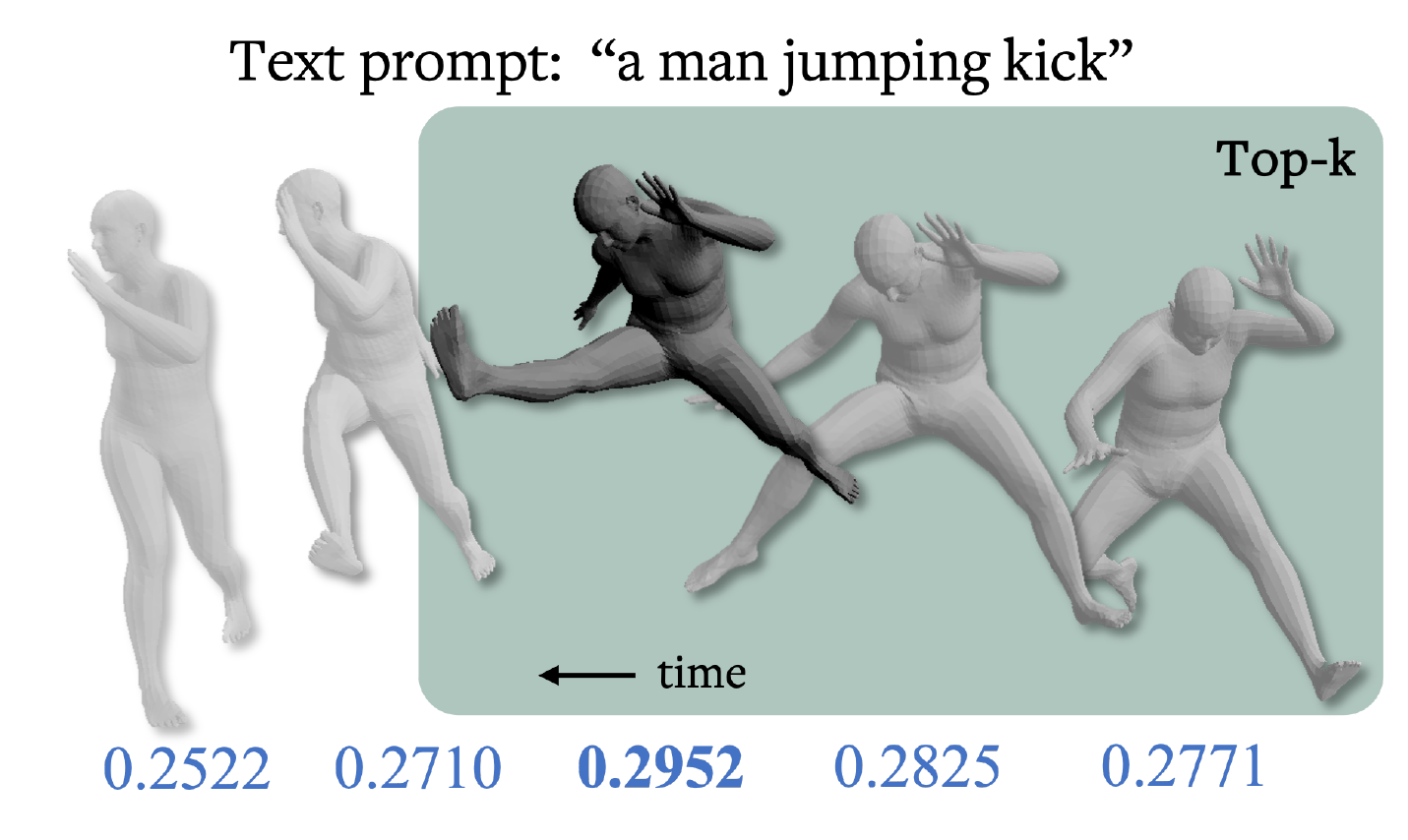}
    \caption{ 
    We choose the top-$\mathtt{k}$ best matching mesh frames that conform with a given text prompt.
    We show the \clipscore{CLIP scores}.}
\label{fig:anchor_sample}
\end{wrapfigure}

However, as careful text prompt and its semantic alignment with the mesh's 
rendered image are crucial for the optimization~\cite{text2mesh,zhang2021pointclip},
we design \emph{multi-modal content mesh sampling} that 
finds the best
text-conforming meshes within the motion.
%
%
Specifically, we render the images of the content mesh sequence, 
$\mathbf{I}(\mathbf{M}_{1:T})$, and compute each image's CLIP similarity score with $y$.
For example, when the text prompt is given as ``a man jumping kick", 
we render each mesh in retrieved motion into an image
and find the semantically matching frames with the jumping kick action
(refer to \Fref{fig:anchor_sample}). 

%


\para{Mask-weighted embedding attention} 
Before the pre-trained CLIP encoder $\mathbf{g(\cdot)}$ encodes 
the rendered images, we apply 
differentiable 2D image augmentations, including random crop and perspective transformations~\cite{ravi2020pytorch3d}.
Such 2D augmentations help
DNSF learn style attributes from 
diverse perspective images, thus can achieve better generalization in 3D contents~\cite{jain2021dreamfields,text2mesh}.

\begin{wrapfigure}[8]{r}{0.4\linewidth}
\centering
    \includegraphics[width=0.9\linewidth]{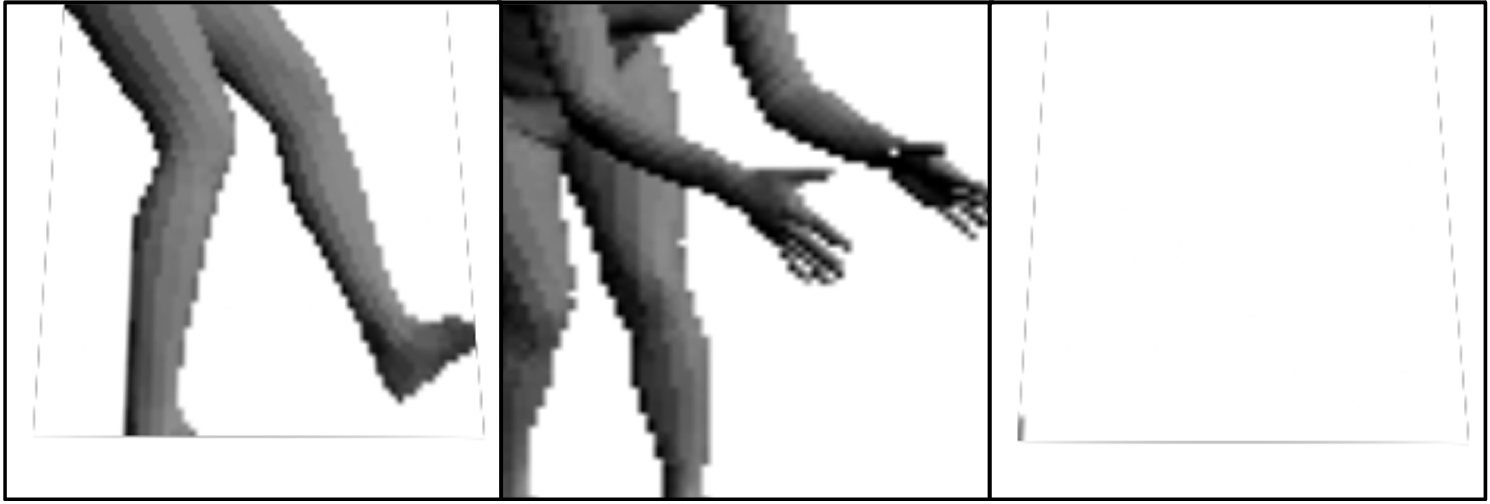}
    \caption{
    Na\"ive mean embedding of the random-cropped renders may distract the optimization of DNSF.}
\label{fig:crop_mesh}
\end{wrapfigure}

However, the problem occurs when careless random crops are applied.
The prior work~\cite{text2mesh} applies extreme close-ups to crop the rendered images,
which severely samples the empty renders
(see \Fref{fig:crop_mesh}).
Such redundant images do not conform to the text prompt
even for the 
properly stylized meshes and distract the stable DNSF optimization.
%

We mitigate this issue by weighting the CLIP embedded vectors $\mathbf{g}(\mathbf{I}_{ij}^{*})$
from $N$ different camera poses $\{\mathbf{p}_j\}_{j=1}^{N}$
according to each image's foreground pixel ratio.
In other words, we reject the embedding vector $\mathbf{g}(\mathbf{I}_{ij}^{*})$,
if $\mathbf{I}_{ij}^{*}$ has an extremely small portion of mesh foreground pixels in it.
We call this \emph{mask-weighted embedding attention}, 
and implement it by simply adding the weight $w_{ij}$ to \Eref{eq:semantic_loss} as:

\begin{equation}
    \mathbf{\bar{g}}(\mathbf{I}^{*}_{i}) = 
    \sum\nolimits_{j=1}^{N} \frac{w_{ij}}{\sum_{k=1}^{N}w_{ik}}
    \mathbf{g}(\mathbf{I}^{*}_{ij}),
    \quad
    w_{ij} = \frac{1}{HW}
    \sum\nolimits_{H,W}
    {\mathbbm{1}[\mathbf{m}_{ij}(h,w) = 1]},
    \label{eq:mask_attention}
\end{equation}
where $H$ and $W$ denote the height and width of the rendered image $\mathbf{I}_{ij}^{*}$ and its foreground mask $\mathbf{m}_{ij}$.


\section{Experiments}
\label{sec:experiment}

In this section, we evaluate CLIP-Actor in several aspects.
Since our model is in an over-fitting regime
and the first approach that addresses the stylization of 3D human meshes in ``motion" 
conditioned on the natural language,
%
we mainly ablate our technical components and the design choices qualitatively.
%
%

Specifically, we describe the models for our evaluation and ablation study in \Sref{sec:model_description}.
We then show our qualitative and quantitative results compared with competing methods in Secs.~\ref{sec:qual} and \ref{sec:quant}.
%
In~\Sref{sec:retrieval_module}, we empirically support our choice of the hierarchical motion retrieval.
%
In \Sref{sec:ablation}, we show how our decoupled style representation,
mesh sampling, view augmentation, and attention mechanism
help us achieve better qualitative results.
%

\begin{figure*}[t]
    \centering
        \includegraphics[width=\linewidth]{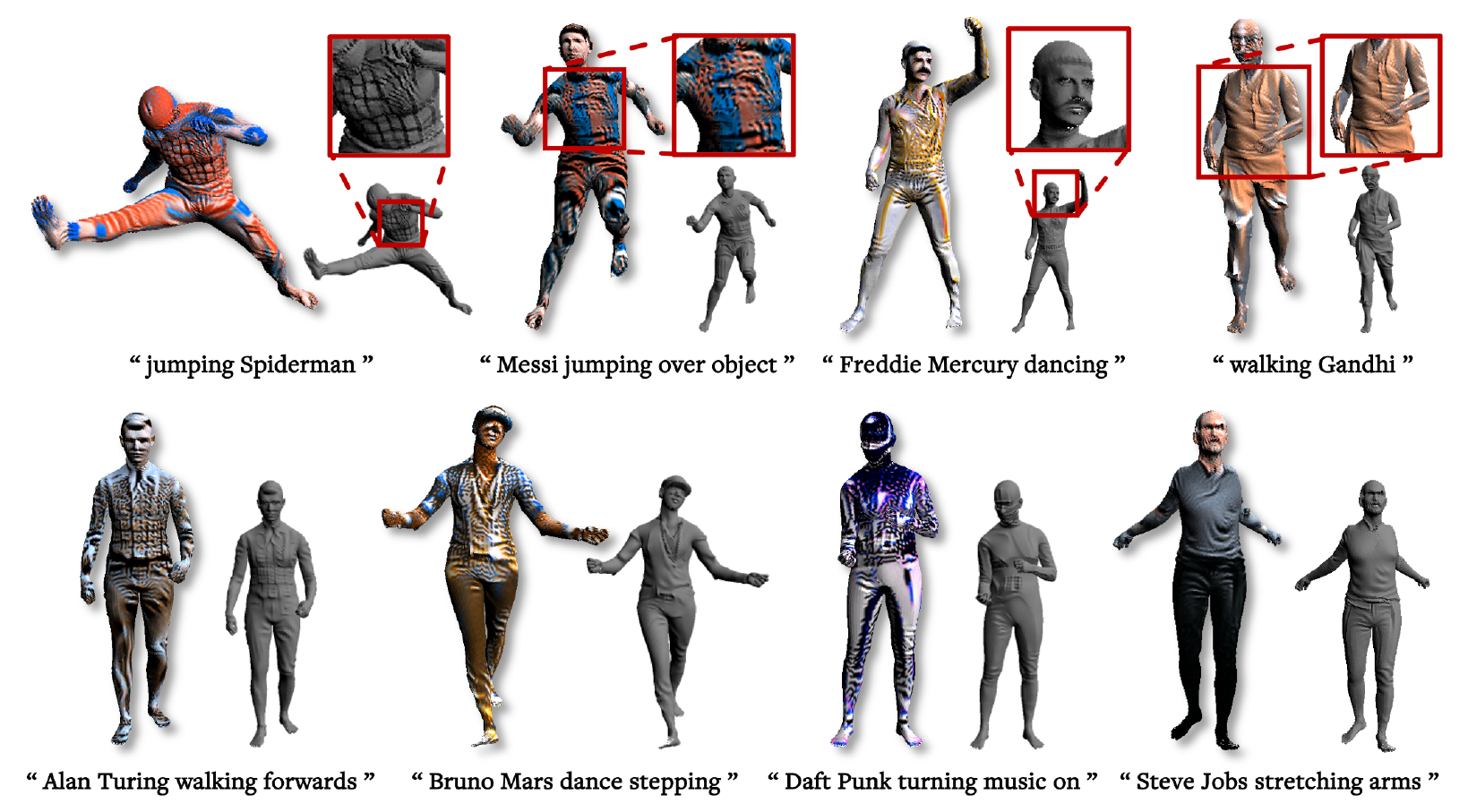}
        \caption{{\textbf{Qualitative results of CLIP-Actor.} 
        Each image shows the representative frame from the recommended motion sequence, with detailed surface geometries and textures, along with the input text prompt. CLIP-Actor shows good action and style consistency, vivid and attractive texture results.
        }}
\label{fig:all_qual}
\end{figure*}

\subsection{Model Description}
\label{sec:model_description}
We define our full model, \textbf{CLIP-Actor}, as the one that uses the top-3 best matching mesh frames 
conforming with a given text prompt, \ie, using multi-modal content mesh sampling and spatio-temporal view augmentations 
along with the mask-weighted embedding attention.
Also, the CLIP-Actor (base) is the model that utilizes only the
center frame of the retrieved motion sequence
and does not utilize DNSF, \ie, using posed mesh to learn the style field. 
Still, CLIP-Actor (base) is a strong baseline model
since it at least mitigates the limitation of Text2Mesh~\cite{text2mesh} 
by suggesting the initial mesh to the neural optimization.


\subsection{Qualitative Results}
\label{sec:qual}
In \Fref{fig:all_qual}, we visualize CLIP-Actor's recommendation
and mesh stylization results for a given text prompt.
%
%
With only a single text prompt, 
CLIP-Actor can retrieve visually conforming motion sequences containing representative poses.
Moreover, CLIP-Actor can capture the subject's representative identities. 
For example, the geometric and texture details such as Spiderman's webbed costume, 
the iconic color of Lionel Messi's uniform, Freddie Mercury's hairstyle, and the robe that Gandhi wears are well-illustrated in \Fref{fig:all_qual}.
%

\begin{figure*}[t]
    \centering
        \includegraphics[width=0.95\linewidth]{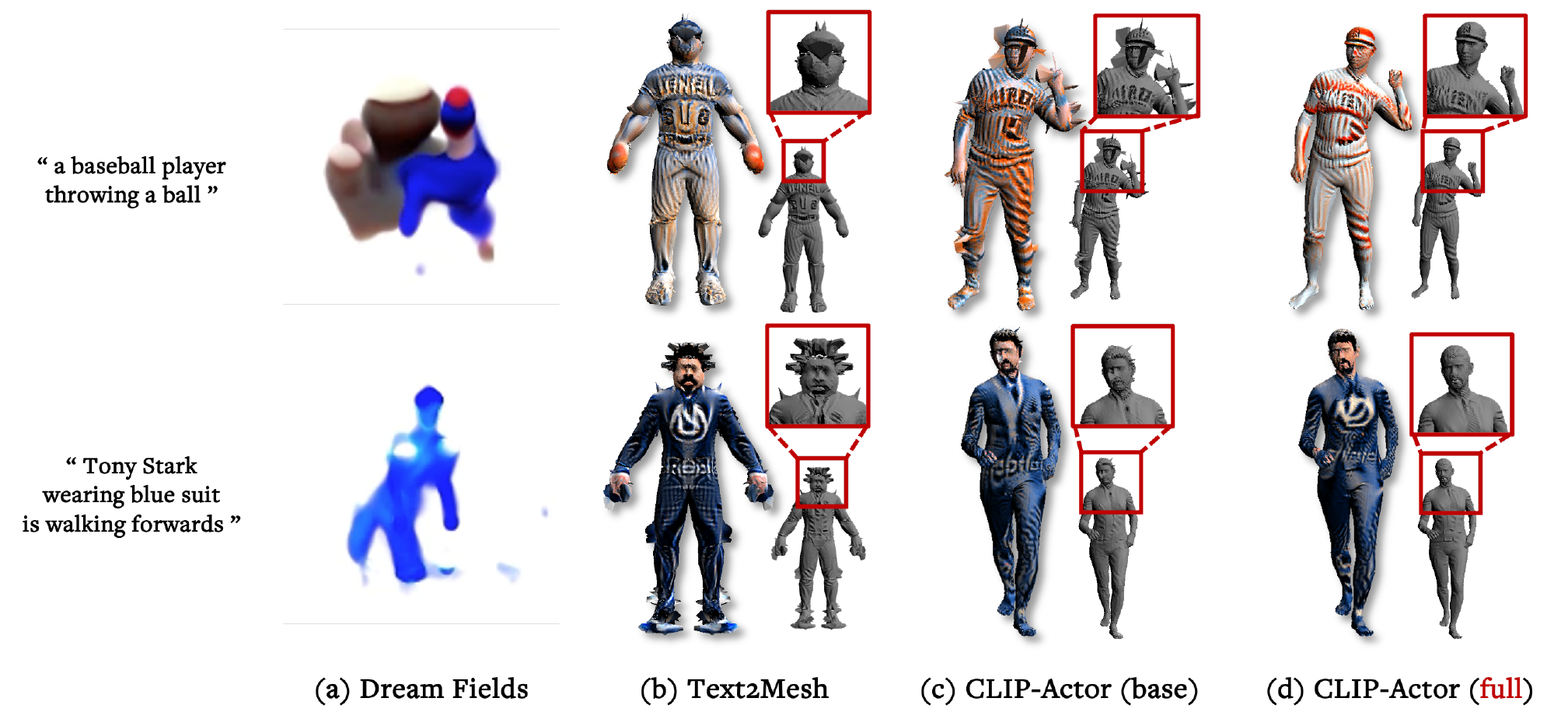}
        \caption{{\textbf{Qualitative comparison.} 
        We compare CLIP-Actor with other competing methods~\cite{jain2021dreamfields,text2mesh}
        and our strong baseline, CLIP-Actor (base). 
        Given the same text prompts as input,
        (a) Dream Fields shows abstract generations, which are blurry and hard to recognize,
        (b) Text2Mesh shows a better generation than Dream Fields but suffers from substantial defects on the surfaces.
        (c) CLIP-Actor (base) shows more text-conforming meshes with human-recognizable
        style attributes but still suffers from surface defects.
        (d) CLIP-Actor shows human-recognizable and semantically conforming action, while presenting
        detailed color and geometry, such as hairstyle and face identities.
        }}
\label{fig:comparison_qual}
\end{figure*}

We also evaluate CLIP-Actor with other recent competing methods, Dream Fields~\cite{jain2021dreamfields} and 
Text2Mesh~\cite{text2mesh}, and our strong baseline model, CLIP-Actor (base).
Figure~\ref{fig:comparison_qual} illustrates the visual comparison of the methods.

Given the same text prompts, Dream Fields shows blurry and non-human-recognizable renderings of the 
generated 3D content.
We postulate that such performance degradation is due to the lack of structural prior 
when training the Dream Fields.
Dream Fields learns the occupancy and the color of 3D points in virtual space
without any structural guidance. 
For example, we cannot impose the human body's physical constraints on Dream Fields
when performing specific actions (refer to \Fref{fig:comparison_qual}\colorref{a}).
We found that applying only semantic supervision to such a highly unrestrained content generation process fails to handle physically constrained human motion and textures.

Text2Mesh shows enhanced texture generation 
than Dream Fields. 
However, it still fails since the given artist-designed human mesh is absolutely uncorrelated
with the target human action. 
Such limitation is originated from Text2Mesh's highly coupled style field, 
which learns the style field from the ``posed" content mesh.
%
Text2Mesh also clamps the per-vertex displacement to lie in a limited range, preventing style attributes from largely changing the content~\cite{text2mesh}. 
On the other hand, 
by adding our novel text-driven human motion recommendation module
before Text2Mesh, and providing
the text-conforming content mesh as an initial point, \ie, CLIP-Actor (base), 
we can significantly enhance Text2Mesh's qualitative performance.
%


Finally, our full CLIP-Actor further enhances the qualitative result by capturing semantically meaningful details such as a cap on a baseball player and the hairstyle of Tony Stark (see \Fref{fig:comparison_qual}\colorref{d}) while reducing the messy spikes.
Our novel DNSF, 
multi-modal content mesh sampling, 
and spatio-temporal view augmentation enable CLIP-Actor to leverage multi-view renderings
originating from multi-frame human motion;
thus, results are much smoother and text-conforming.
More importantly, note that all the other methods except CLIP-Actor
cannot handle human motion. CLIP-Actor recommends text-conforming human motion
and synthesizes temporally consistent and pose-agnostic mesh style attributes.
%

\subsection{Quantitative Results} \label{sec:quant}

\begin{wrapfigure}[10]{r}{0.44\linewidth}
\centering
    \includegraphics[width=\linewidth]{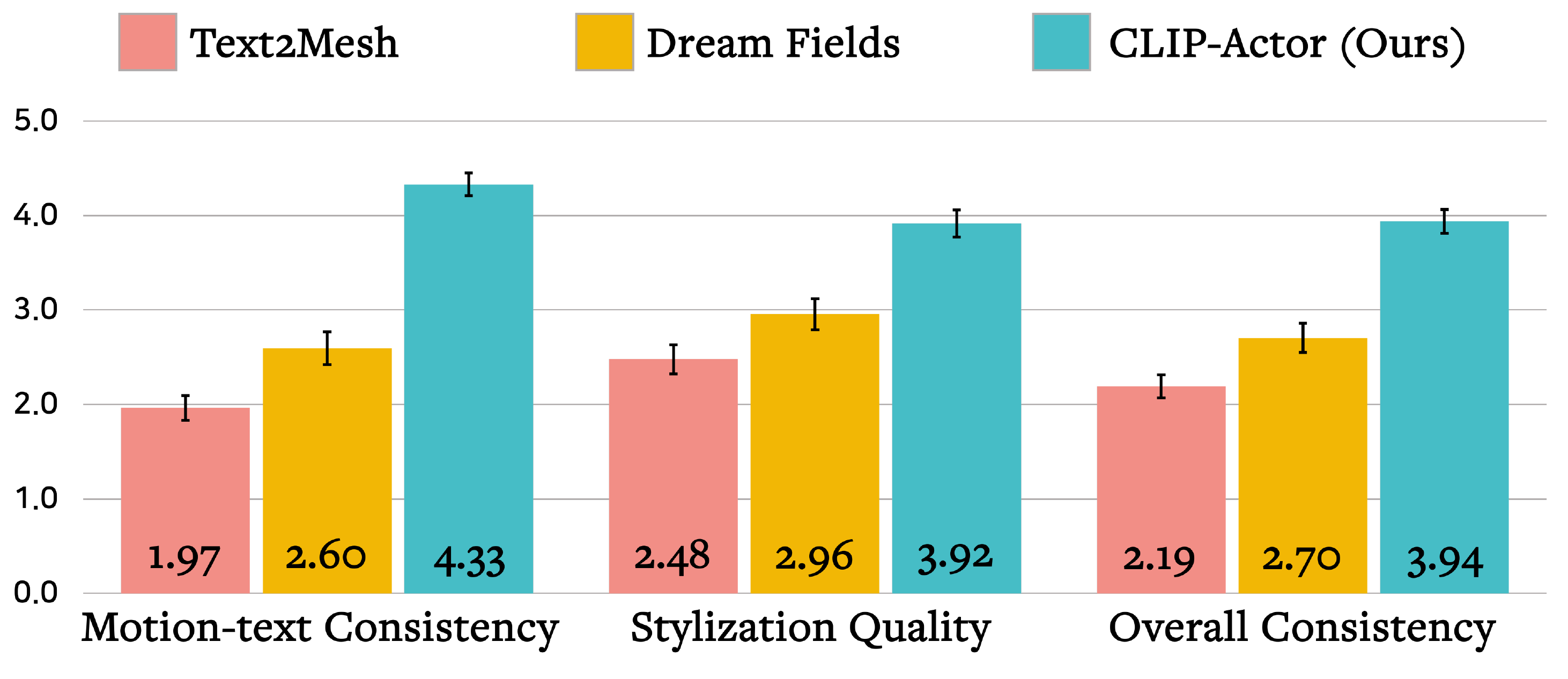}
    \caption{{\textbf{User study results.} 
   CLIP-Actor outperforms other competing methods in various aspects.}}
\label{fig:userstudy}
\end{wrapfigure}
Since there are no benchmarks for our task, we conduct a user study to evaluate CLIP-Actor quantitatively.
%
We ask 46 non-expert users to score (1-5) five random text-avatar paired results
regarding motion-text consistency, stylization quality, and overall consistency with the given text. 
Figure \ref{fig:userstudy} shows that our CLIP-Actor outperforms other competing methods in all aspects,
while none of the competing methods scored higher than a neutral point (\ie, ${<}3$).
The differences are significantly noticeable in the motion-text consistency, which validates our good action consistency.

\subsection{Evaluation on Retrieval Module}
\label{sec:retrieval_module}
We validate the performance of our retrieval module design choice 
by comparing it with other variants.
%
We use the SICK dataset~\cite{Marelli2014ASC} that contains
contextually similar sentence pairs generated from image descriptions.
We build module variants to simulate various retrieval scenarios. 
Details about the dataset and the experiment settings can be found in the supplementary material.

\para{Retrieval module variants}
We consider two hierarchical modules and two single-stage baselines 
for the design variants.
Our full hierarchical retrieval module
utilizes 
pre-trained CLIP and MPNet sequentially (see CLIP+MPNet in \Sref{sec:4_motion_recommendation}).
We also design the reverse ordered hierarchical module, MPNet+CLIP.
For hierarchical models, 
top-$\textrm{k}$ candidates are matched 
at the first stage,
and the best-matched item is selected after re-ranking.
Finally, our single-stage baselines are the modules that only use either pre-trained CLIP or MPNet encoder.

\para{Quantitative results}
We empirically show the performance of our hierarchical motion retrieval module by comparing it with the different design variants.
%
As shown in \Tref{table:retrieval_precison}\colorref{a}, the single-stage baselines
show comparable results with our hierarchical model in the SICK4.8 setting, where the sentence pairs are more related to each other (refer to \Tref{table:retrieval_precison}\colorref{b}).
However, CLIP shows higher precision than MPNet in SICK4.4.
We postulate that CLIP catches the visual semantics, 
\eg, similar context can be imagined from ``performing with a guitar" and ``playing a guitar".
On the other hand, the language expert, MPNet, focuses on the textual difference of description,
\eg, ``performing with" and ``playing". Thus, it is 
sensitive to the 
textual structure. 
%
In the setting with the increased number of samples, \ie, SICK[4.4,4.8],
CLIP shows comparable results with ours but is still insufficient without the help of the language expert.
%
Moreover, the MPNet+CLIP shows unstable performances over settings. 
In contrast, our full retrieval module
consistently outperforms over all settings.
We demonstrate that a coarse-to-fine matching system achieves favorable retrieval performance on natural language.


\begin{table}[t]

\caption{\textbf{Evaluation results on the retrieval module}. (a) The precision is measured as the percentage of matched sentence pairs among all pairs. (b) The samples of each range of the SICK dataset. Our hierarchical retrieval outperforms all the other variants.}
    \centering
    \begin{tabular}{c c c}
        \small{(a)} && \small{(b)}\\
        \resizebox{0.59\linewidth}{!}{%
            \begin{tabular}{ l c c c c c } 
                \toprule
                    & \multicolumn{5}{ c }{{\bfseries Retrieval Precision [\%]}} \\ 
                    \cmidrule{2-6}
                    Retrieval Module & \small{SICK4.8} && \small{SICK4.4} && \small{SICK[4.4,4.8]} \\
                    \midrule
                    CLIP & 91.94 && 85.21 && 81.62 \\
                    MPNet & 91.94 && 83.56 && 80.55 \\
                    MPNet+CLIP & 91.34 && 85.48 && 80.41 \\
                    \cmidrule(r){1-6}
                    CLIP+MPNet (Ours) & \textbf{92.24} && \textbf{85.75} && \textbf{81.90} \\
                \bottomrule
            \end{tabular}
        }&&
        \raisebox{-0.48\height }{\includegraphics[width=0.4\linewidth]{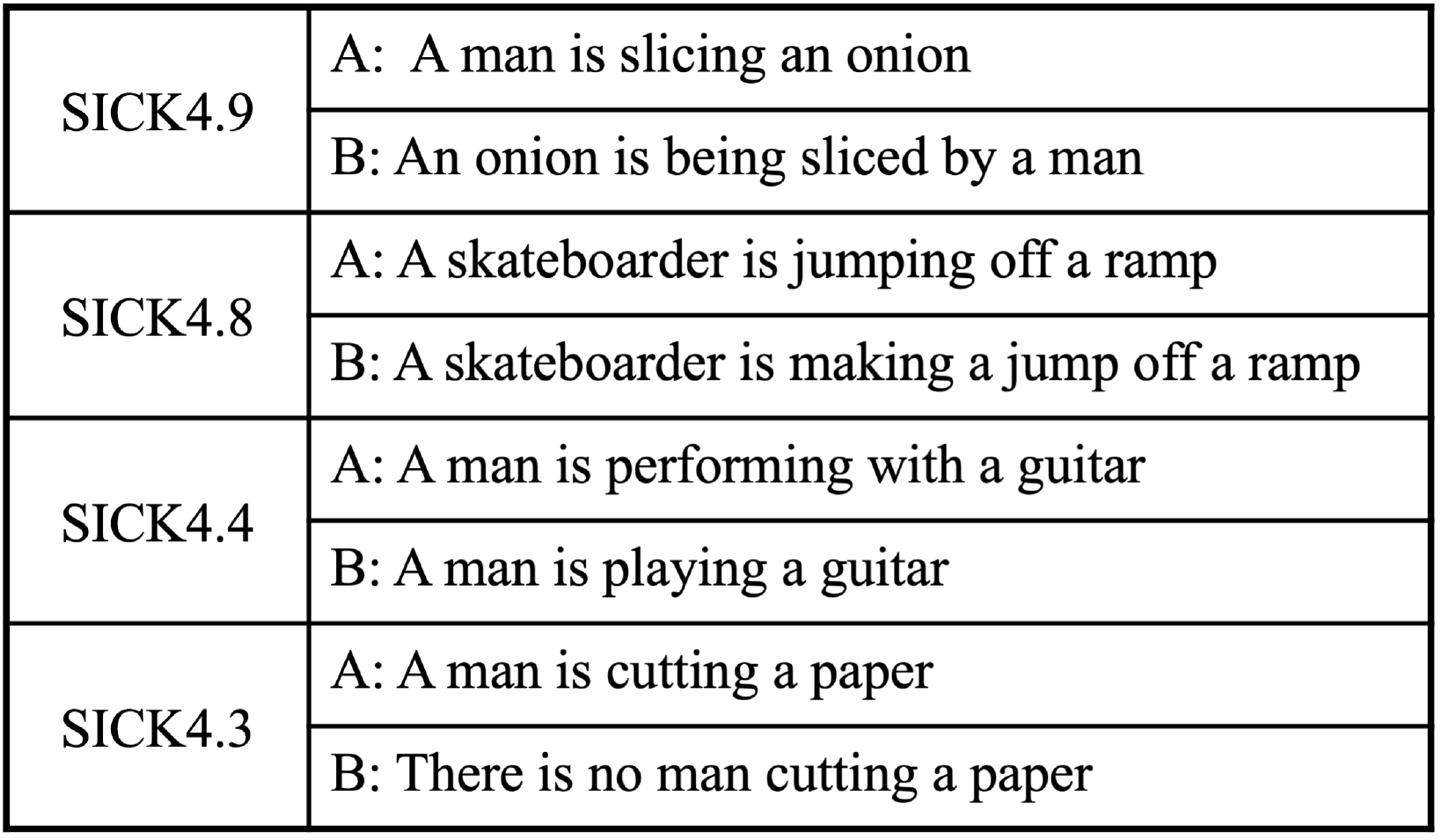}}
    \end{tabular}
\label{table:retrieval_precison}
\end{table}

\subsection{Ablation on Decoupled Neural Style Fields (DNSF)}
\label{sec:ablation}
\begin{figure*}[t]
\centering
    \includegraphics[width=0.8\linewidth]{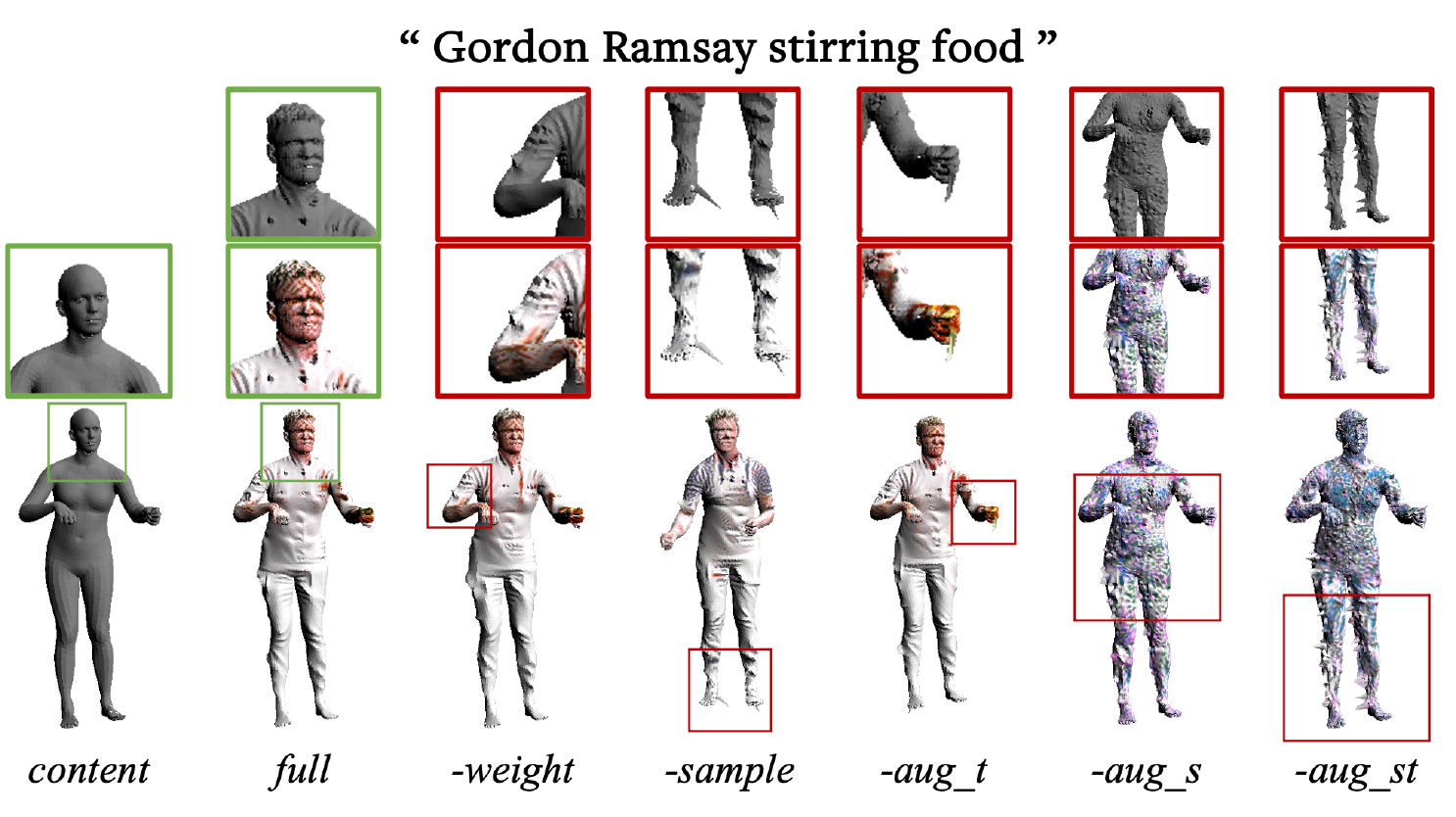}
    \caption{{\textbf{Ablation results.} 
    We remove each of CLIP-Actor's components to validate corresponding effects.
    Our \textit{full} model shows the most smooth geometry and vivid color.}}
\label{fig:ablation}
\end{figure*}

We analyze CLIP-Actor by ablating each of the components of DNSF.
Figure~\ref{fig:ablation} shows the qualitative ablation results for our major technical components. 

\para{Effects of temporal augmentation}
First, we remove
temporal view augmentation 
so that DNSF utilizes only a single mesh frame (top-1).
Removing the multi-frame renderings significantly degrades 
the visual quality, where it presents noticeable spikes on the surface and unrealistic colors ($-\textit{aug}\_\textit{t}$ in~\Fref{fig:ablation}).
Since our full model utilizes top-3 relevant frames and 
2D, 3D augmentations,
it leverages multi-view of stylized mesh, which regularizes the model from overfitting~\cite{text2mesh,jain2021dreamfields}. 

\para{Effects of multi-modal content mesh sampling}
We also compare the full CLIP-Actor with the model without \emph{multi-modal content mesh sampling}.
Multi-modal content mesh sampling enables DNSF to begin its optimization
with better initialization that conforms with the text prompt. 
Na\"ive sampling of the content mesh 
yields unrecognizable 
face identities, degraded texture, and geometric details 
(see $-\textit{sample}$ in~\Fref{fig:ablation}).
\para{Effects of mask-weighted embedding attention}
The \emph{mask-weighted embedding attention}
adds detailed touches to the stylized meshes.
By preventing 
empty renderings
from guiding the optimization, 
%
%
%
it enables learning fine geometric and texture details 
via focused gradient flow in back-propagation.
When the augmented rendered images contain extreme close-ups of distal body regions, 
such as tiptoe or fingertips, 
our embedding attention method draws the DNSF's attention to the mesh foreground pixels 
rather than empty space with focused gradient flow.
In \Fref{fig:ablation}, $\textit{-weight}$ shows the result when we train DNSF without 
our attention mechanism.
Our full CLIP-Actor shows much smooth, fine-grained geometric details.
We believe our novel attention mechanism can be applied to not only text-driven 3D object manipulation pipelines~\cite{text2mesh,jain2021dreamfields,zhang2021pointclip} but also differentiable rendering applications~\cite{mildenhall2020nerf,Jain_2021_ICCV,yu2020pixelnerf}.


\section{Conclusion}
We present CLIP-Actor, a text-driven animated human mesh synthesis system.
%
Leveraging multi-modal aware and semantic textual matching, CLIP-Actor recommends the best semantically matching human motion sequence with the input text prompt
in a hierarchical manner. 
Our CLIP-Actor then stylizes the meshes of recommended motion by 
synthesis-through-optimization in a pose-agnostic manner
via decoupled neural style fields. 
%
We further develop novel neural optimization techniques to 
utilize multi-modal sampling and embedding weighting, which 
stabilize and enhance the detailization and texturization quality.
%

CLIP-Actor can be extended to 
other parametric mesh models,
such as hands 
and animals~\cite{MANO:SIGGRAPHASIA:2017,Zuffi:CVPR:2017,biggs2020wldo,biggs2018creatures,youwang2021unified,threedsafari,face3dmm}, 
enabling diverse animation of 3D objects. 
%
%
One promising future application of CLIP-Actor would be a dataset generation of 
stylized meshes in motion, paired with natural language description. 
We believe such multi-modal datasets can boost exciting future applications.

\vfill

\para{Acknowledgment}
This work was supported by Institute of Information \& communications Technology Planning \& Evaluation (IITP) grant funded by the Korea government(MSIT) (No.2022-00164860, Development of Human Digital Twin Technology Based on Dynamic Behavior Modeling and Human-Object-Space Interaction; and No.2021-0-02068, Artificial Intelligence Innovation Hub). 
\clearpage

\bibliographystyle{splncs04}
\bibliography{egbib}

\begin{thebibliography}{10}
\providecommand{\url}[1]{\texttt{#1}}
\providecommand{\urlprefix}{URL }
\providecommand{\doi}[1]{https://doi.org/#1}

\bibitem{Agirre2012SemEval2012T6}
Agirre, E., Diab, M., Cer, D., Gonzalez-Agirre, A.: Semeval-2012 task 6: A
  pilot on semantic textual similarity. In: Proceedings of the First Joint
  Conference on Lexical and Computational Semantics - Volume 1: Proceedings of
  the Main Conference and the Shared Task, and Volume 2: Proceedings of the
  Sixth International Workshop on Semantic Evaluation. p. 385–393 (2012)

\bibitem{ahn:Text2Action:icra18}
Ahn, H., Ha, T., Choi, Y., Yoo, H., Oh, S.: Text2action: Generative adversarial
  synthesis from language to action. In: IEEE International Conference on
  Robotics and Automation (ICRA) (2018)

\bibitem{Ahuja2019Language2PoseNL}
Ahuja, C., Morency, L.P.: Language2pose: Natural language grounded pose
  forecasting. International Conference on 3D Vision (3DV)  (2019)

\bibitem{barron2021mipnerf360}
Barron, J.T., Mildenhall, B., Verbin, D., Srinivasan, P.P., Hedman, P.:
  Mip-nerf 360: Unbounded anti-aliased neural radiance fields. IEEE Conference
  on Computer Vision and Pattern Recognition (CVPR)  (2021)

\bibitem{bhatnagar2020ipnet}
Bhatnagar, B.L., Sminchisescu, C., Theobalt, C., Pons-Moll, G.: Combining
  implicit function learning and parametric models for 3d human reconstruction.
  In: European Conference on Computer Vision (ECCV) (2020)

\bibitem{Bhatnagar2019MultiGarmentNL}
Bhatnagar, B.L., Tiwari, G., Theobalt, C., Pons-Moll, G.: Multi-garment net:
  Learning to dress 3d people from images. In: IEEE International Conference on
  Computer Vision (ICCV) (2019)

\bibitem{biggs2020wldo}
Biggs, B., Boyne, O., Charles, J., Fitzgibbon, A., Cipolla, R.: {W}ho left the
  dogs out?: {3D} animal reconstruction with expectation maximization in the
  loop. In: European Conference on Computer Vision (ECCV) (2020)

\bibitem{biggs2018creatures}
Biggs, B., Roddick, T., Fitzgibbon, A., Cipolla, R.: {C}reatures great and
  {SMAL}: {R}ecovering the shape and motion of animals from video. In: Asia
  Conference on Computer Vision (ACCV) (2018)

\bibitem{face3dmm}
Blanz, V., Vetter, T.: A morphable model for the synthesis of 3d faces. In:
  SIGGRAPH (1999)

\bibitem{bozic2021neuraldeformationgraphs}
Bo{\v{z}}i{\v{c}}, A., Palafox, P., Zollh{\"o}fer, M., Thies, J., Dai, A.,
  Nie{\ss}ner, M.: Neural deformation graphs for globally-consistent non-rigid
  reconstruction. In: Advances in Neural Information Processing Systems
  (NeurIPS) (2021)

\bibitem{bozic2020neuraltracking}
Bozic, A., Palafox, P., Zoll{\"o}fer, M., Dai, A., Thies, J., Nie{\ss}ner, M.:
  Neural non-rigid tracking. In: Advances in Neural Information Processing
  Systems (NeurIPS) (2020)

\bibitem{Burov2021DynamicSF}
Burov, A., Nie{\ss}ner, M., Thies, J.: Dynamic surface function networks for
  clothed human bodies. In: IEEE International Conference on Computer Vision
  (ICCV) (2021)

\bibitem{Canfes2022TextAI}
Canfes, Z., Atasoy, M.F., Dirik, A., Yanardag, P.: Text and image guided 3d
  avatar generation and manipulation. arXiv:2202.06079  (2022)

\bibitem{rhoi2020}
Cao, Z., Radosavovic, I., Kanazawa, A., Malik, J.: Reconstructing hand-object
  interactions in the wild. IEEE International Conference on Computer Vision
  (ICCV)  (2021)

\bibitem{du2021gem}
Du, Y., Collins, M.K., Tenenbaum, B.J., Sitzmann, V.: Learning signal-agnostic
  manifolds of neural fields. In: Advances in Neural Information Processing
  Systems (NeurIPS) (2021)

\bibitem{recurrent_katerina}
Fragkiadaki, K., Levine, S., Felsen, P., Malik, J.: Recurrent network models
  for human dynamics. IEEE International Conference on Computer Vision (ICCV)
  (2015)

\bibitem{Frans2021CLIPDrawET}
Frans, K., Soros, L.B., Witkowski, O.: Clipdraw: Exploring text-to-drawing
  synthesis through language-image encoders. arXiv:2106.14843  (2021)

\bibitem{Gafni_2021_CVPR}
Gafni, G., Thies, J., Zollh{\"o}fer, M., Nie{\ss}ner, M.: Dynamic neural
  radiance fields for monocular 4d facial avatar reconstruction. In: IEEE
  Conference on Computer Vision and Pattern Recognition (CVPR) (2021)

\bibitem{Gal2021StyleGANNADACD}
Gal, R., Patashnik, O., Maron, H., Chechik, G., Cohen-Or, D.: Stylegan-nada:
  Clip-guided domain adaptation of image generators. In: ACM Transactions on
  Graphics (SIGGRAPH) (2022)

\bibitem{Ghosh2021SynthesisOC}
Ghosh, A., Cheema, N., Oguz, C., Theobalt, C., Slusallek, P.: Synthesis of
  compositional animations from textual descriptions. IEEE International
  Conference on Computer Vision (ICCV)  (2021)

\bibitem{guo2022action2video}
Guo, C., Zuo, X., Wang, S., Liu, X., Zou, S., Gong, M., Cheng, L.:
  Action2video: Generating videos of human 3d actions. International Journal of
  Computer Vision (IJCV) pp. 1--31 (2022)

\bibitem{chuan2020action2motion}
Guo, C., Zuo, X., Wang, S., Zou, S., Sun, Q., Deng, A., Gong, M., Cheng, L.:
  Action2motion: Conditioned generation of 3d human motions. In: ACM
  International Conference on Multimedia (MM) (2020)

\bibitem{inversesim}
Guo, J., Li, J., Narain, R., Park, H.: Inverse simulation: Reconstructing
  dynamic geometry of clothed humans via optimal control. In: IEEE Conference
  on Computer Vision and Pattern Recognition (CVPR) (2021)

\bibitem{Hanser2009SceneMakerIM}
Hanser, E., Kevitt, P.M., Lunney, T.F., Condell, J.: Scenemaker: Intelligent
  multimodal visualisation of natural language scripts. In: Proceedings of the
  20th Irish conference on Artificial intelligence and cognitive science (2009)

\bibitem{Hodosh2013FramingID}
Hodosh, M., Young, P., Hockenmaier, J.: Framing image description as a ranking
  task: Data, models and evaluation metrics. Journal of Artificial Intelligence
  Research  \textbf{47},  853--899 (2013)

\bibitem{jain2021dreamfields}
Jain, A., Mildenhall, B., Barron, J.T., Abbeel, P., Poole, B.: Zero-shot
  text-guided object generation with dream fields. In: IEEE Conference on
  Computer Vision and Pattern Recognition (CVPR) (2022)

\bibitem{Jain_2021_ICCV}
Jain, A., Tancik, M., Abbeel, P.: Putting nerf on a diet: Semantically
  consistent few-shot view synthesis. In: IEEE International Conference on
  Computer Vision (ICCV) (2021)

\bibitem{trans_vae}
Jiang, J., Xia, G.G., Carlton, D.B., Anderson, C.N., Miyakawa, R.H.:
  Transformer vae: A hierarchical model for structure-aware and interpretable
  music representation learning. In: IEEE International Conference on
  Acoustics, Speech, and Signal Processing (ICASSP) (2020)

\bibitem{kato2020differentiable}
Kato, H., Beker, D., Morariu, M., Ando, T., Matsuoka, T., Kehl, W., Gaidon, A.:
  Differentiable rendering: A survey. arXiv:2006.12057  (2020)

\bibitem{Kim2021DiffusionCLIPTD}
Kim, G., Ye, J.C.: Diffusionclip: Text-guided diffusion models for robust image
  manipulation. In: IEEE Conference on Computer Vision and Pattern Recognition
  (CVPR) (2022)

\bibitem{Kwon2021CLIPstylerIS}
Kwon, G., Ye, J.C.: Clipstyler: Image style transfer with a single text
  condition. In: IEEE Conference on Computer Vision and Pattern Recognition
  (CVPR) (2022)

\bibitem{Li2021NClothP3}
Li, Y., Tang, M., Yang, Y., Huang, Z., Tong, R., Yang, S., Li, Y., Manocha, D.:
  {N-Cloth}: Predicting {3D} cloth deformation with mesh-based networks.
  Computer Graphics Forum (Proceedings of Eurographics) pp. 547--558 (2022)

\bibitem{lin:vigil18}
Lin, A.S., Wu, L., Corona, R., Tai, K.W.H., Huang, Q., Mooney, R.J.: Generating
  animated videos of human activities from natural language descriptions. In:
  Proceedings of the Visually Grounded Interaction and Language Workshop at
  NeurIPS 2018 (2018)

\bibitem{SMPL:2015}
Loper, M., Mahmood, N., Romero, J., Pons-Moll, G., Black, M.J.: {SMPL}: A
  skinned multi-person linear model. ACM Transactions on Graphics (SIGGRAPH
  Asia)  \textbf{34}(6), ~248 (2015)

\bibitem{ma2020cape}
Ma, Q., Yang, J., Ranjan, A., Pujades, S., Pons-Moll, G., Tang, S., Black,
  M.J.: Learning to dress 3d people in generative clothing. In: IEEE Conference
  on Computer Vision and Pattern Recognition (CVPR) (2020)

\bibitem{AMASS:ICCV:2019}
Mahmood, N., Ghorbani, N., Troje, N.F., Pons-Moll, G., Black, M.J.: {AMASS}:
  Archive of motion capture as surface shapes. In: IEEE International
  Conference on Computer Vision (ICCV) (2019)

\bibitem{Marelli2014ASC}
Marelli, M., Menini, S., Baroni, M., Bentivogli, L., Bernardi, R., Zamparelli,
  R.: A sick cure for the evaluation of compositional distributional semantic
  models. In: International Conference on Language Resources and Evaluation
  (LREC) (2014)

\bibitem{text2mesh}
Michel, O., Bar-On, R., Liu, R., Benaim, S., Hanocka, R.: Text2mesh:
  Text-driven neural stylization for meshes. In: IEEE Conference on Computer
  Vision and Pattern Recognition (CVPR) (2022)

\bibitem{mildenhall2020nerf}
Mildenhall, B., Srinivasan, P.P., Tancik, M., Barron, J.T., Ramamoorthi, R.,
  Ng, R.: Nerf: Representing scenes as neural radiance fields for view
  synthesis. In: European Conference on Computer Vision (ECCV) (2020)

\bibitem{Nichol2021GLIDETP}
Nichol, A., Dhariwal, P., Ramesh, A., Shyam, P., Mishkin, P., McGrew, B.,
  Sutskever, I., Chen, M.: Glide: Towards photorealistic image generation and
  editing with text-guided diffusion models. In: International Conference on
  Machine Learning (ICML) (2022)

\bibitem{Niemeyer2020CVPR}
Niemeyer, M., Mescheder, L., Oechsle, M., Geiger, A.: Differentiable volumetric
  rendering: Learning implicit 3d representations without 3d supervision. In:
  IEEE Conference on Computer Vision and Pattern Recognition (CVPR) (2020)

\bibitem{palafox2021npm}
Palafox, P., Bozic, A., Thies, J., Nie{\ss}ner, M., Dai, A.: Neural parametric
  models for 3d deformable shapes. In: IEEE International Conference on
  Computer Vision (ICCV) (2021)

\bibitem{Patashnik_2021_ICCV}
Patashnik, O., Wu, Z., Shechtman, E., Cohen-Or, D., Lischinski, D.: Styleclip:
  Text-driven manipulation of stylegan imagery. In: IEEE International
  Conference on Computer Vision (ICCV) (2021)

\bibitem{SMPL-X:2019}
Pavlakos, G., Choutas, V., Ghorbani, N., Bolkart, T., Osman, A.A.A., Tzionas,
  D., Black, M.J.: Expressive body capture: {3D} hands, face, and body from a
  single image. In: IEEE Conference on Computer Vision and Pattern Recognition
  (CVPR) (2019)

\bibitem{petrovich21actor}
Petrovich, M., Black, M.J., Varol, G.: Action-conditioned 3{D} human motion
  synthesis with transformer {VAE}. In: IEEE International Conference on
  Computer Vision (ICCV) (2021)

\bibitem{petrovich22temos}
Petrovich, M., Black, M.J., Varol, G.: {TEMOS}: Generating diverse human
  motions from textual descriptions. In: European Conference on Computer Vision
  ({ECCV}) (2022)

\bibitem{Plappert2018LearningAB}
Plappert, M., Mandery, C., Asfour, T.: Learning a bidirectional mapping between
  human whole-body motion and natural language using deep recurrent neural
  networks. Robotics and Autonomous Systems  \textbf{109},  13--26 (2018)

\bibitem{BABEL:CVPR:2021}
Punnakkal, A.R., Chandrasekaran, A., Athanasiou, N., Quiros-Ramirez, A., Black,
  M.J.: {BABEL}: Bodies, action and behavior with english labels. In: IEEE
  Conference on Computer Vision and Pattern Recognition (CVPR) (2021)

\bibitem{radford2021learning}
Radford, A., Kim, J.W., Hallacy, C., Ramesh, A., Goh, G., Agarwal, S., Sastry,
  G., Askell, A., Mishkin, P., Clark, J., Krueger, G., Sutskever, I.: Learning
  transferable visual models from natural language supervision. In:
  International Conference on Machine Learning (ICML) (2021)

\bibitem{ravi2020pytorch3d}
Ravi, N., Reizenstein, J., Novotny, D., Gordon, T., Lo, W.Y., Johnson, J.,
  Gkioxari, G.: Accelerating 3d deep learning with pytorch3d. arXiv:2007.08501
  (2020)

\bibitem{MANO:SIGGRAPHASIA:2017}
Romero, J., Tzionas, D., Black, M.J.: Embodied hands: Modeling and capturing
  hands and bodies together. ACM Transactions on Graphics (SIGGRAPH Asia)
  \textbf{36}(6) (Nov 2017)

\bibitem{saito2019pifu}
Saito, S., Huang, Z., Natsume, R., Morishima, S., Kanazawa, A., Li, H.: Pifu:
  Pixel-aligned implicit function for high-resolution clothed human
  digitization. In: IEEE International Conference on Computer Vision (ICCV)
  (2019)

\bibitem{saito2020pifuhd}
Saito, S., Simon, T., Saragih, J., Joo, H.: Pifuhd: Multi-level pixel-aligned
  implicit function for high-resolution 3d human digitization. In: IEEE
  Conference on Computer Vision and Pattern Recognition (CVPR) (2020)

\bibitem{Saito:CVPR:2021}
Saito, S., Yang, J., Ma, Q., Black, M.J.: {SCANimate}: Weakly supervised
  learning of skinned clothed avatar networks. In: IEEE Conference on Computer
  Vision and Pattern Recognition (CVPR) (2021)

\bibitem{Sanh2019DistilBERTAD}
Sanh, V., Debut, L., Chaumond, J., Wolf, T.: Distilbert, a distilled version of
  bert: smaller, faster, cheaper and lighter. arXiv:1910.01108  (2019)

\bibitem{shree2021exploiting}
Shree, V., Asfora, B., Zheng, R., Hong, S., Banfi, J., Campbell, M.: Exploiting
  natural language for efficient risk-aware multi-robot sar planning. IEEE
  Robotics and Automation Letters  \textbf{6}(2),  3152--3159 (2021)

\bibitem{song2020mpnet}
Song, K., Tan, X., Qin, T., Lu, J., Liu, T.Y.: Mpnet: Masked and permuted
  pre-training for language understanding. Advances in Neural Information
  Processing Systems (NeurIPS)  (2020)

\bibitem{tancik2020fourfeat}
Tancik, M., Srinivasan, P.P., Mildenhall, B., Fridovich-Keil, S., Raghavan, N.,
  Singhal, U., Ramamoorthi, R., Barron, J.T., Ng, R.: Fourier features let
  networks learn high frequency functions in low dimensional domains. Advances
  in Neural Information Processing Systems (NeurIPS)  (2020)

\bibitem{tevet2022motionclip}
Tevet, G., Gordon, B., Hertz, A., Bermano, A.H., Cohen-Or, D.: Motionclip:
  Exposing human motion generation to clip space. In: European Conference on
  Computer Vision (ECCV) (2022)

\bibitem{Kaolin2019}
Tsang, C.F., Shugrina, M., Lafleche, J.F., Takikawa, T., Wang, J., Loop, C.,
  Chen, W., Jatavallabhula, K.M., Smith, E., Rozantsev, A., Perel, O., Shen,
  F., Gao, J., Fidler, S., State, G., Gorski, J., Xiang, T., Li, J., Li, M.,
  Lebaredian, R.: Kaolin (2019)

\bibitem{wang2021clip}
Wang, C., Chai, M., He, M., Chen, D., Liao, J.: Clip-nerf: Text-and-image
  driven manipulation of neural radiance fields. In: IEEE Conference on
  Computer Vision and Pattern Recognition (CVPR) (2022)

\bibitem{Wang2021NEURIPS}
Wang, S., Mihajlovic, M., Ma, Q., Geiger, A., Tang, S.: Metaavatar: Learning
  animatable clothed human models from few depth images. In: Advances in Neural
  Information Processing Systems (NeurIPS) (2021)

\bibitem{yoonICRA19}
Yoon, Y., Ko, W.R., Jang, M., Lee, J., Kim, J., Lee, G.: Robots learn social
  skills: End-to-end learning of co-speech gesture generation for humanoid
  robots. In: IEEE International Conference on Robotics and Automation (ICRA)
  (2019)

\bibitem{youwang2021unified}
Youwang, K., Ji-Yeon, K., Joo, K., Oh, T.H.: Unified 3d mesh recovery of humans
  and animals by learning animal exercise. In: British Machine Vision
  Conference (BMVC) (2021)

\bibitem{yu2020pixelnerf}
Yu, A., Ye, V., Tancik, M., Kanazawa, A.: {pixelNeRF}: Neural radiance fields
  from one or few images. In: IEEE Conference on Computer Vision and Pattern
  Recognition (CVPR) (2021)

\bibitem{zhang2020phosa}
Zhang, J.Y., Pepose, S., Joo, H., Ramanan, D., Malik, J., Kanazawa, A.:
  Perceiving 3d human-object spatial arrangements from a single image in the
  wild. In: European Conference on Computer Vision (ECCV) (2020)

\bibitem{zhang2021pointclip}
Zhang, R., Guo, Z., Zhang, W., Li, K., Miao, X., Cui, B., Qiao, Y., Gao, P.,
  Li, H.: Pointclip: Point cloud understanding by clip. In: IEEE Conference on
  Computer Vision and Pattern Recognition (CVPR) (2021)

\bibitem{threedsafari}
Zuffi, S., Kanazawa, A., Berger-Wolf, T., Black, M.J.: Three-d safari: Learning
  to estimate zebra pose, shape, and texture from images “in the wild”. In:
  IEEE International Conference on Computer Vision (ICCV) (2019)

\bibitem{Zuffi:CVPR:2017}
Zuffi, S., Kanazawa, A., Jacobs, D., Black, M.J.: {3D} menagerie: Modeling the
  {3D} shape and pose of animals. In: IEEE Conference on Computer Vision and
  Pattern Recognition (CVPR) (2017)

\end{thebibliography}
\renewcommand\thesection{\Alph{section}}
\renewcommand\thefigure{\alph{figure}}
\setcounter{section}{0}
\setcounter{figure}{0}

\setlength{\intextsep}{0.5mm}

\title{
CLIP-Actor: Text-Driven 
Recommendation \\ and Stylization 
for Animating Human Meshes\\
\textmd{--- Supplementary Material ---}}
\titlerunning{CLIP-Actor}
%
\author{
Kim Youwang\inst{1}\thanks{Authors contributed equally to this work.}
\qquad
Kim Ji-Yeon\inst{2}\samethanks
\qquad
Tae-Hyun Oh\inst{1,2}\thanks{Joint affiliated with Yonsei University, Korea.}
}
\authorrunning{Youwang et al.}

\institute{
Department of ${}^1$EE \& ${}^2$CiTE, POSTECH\\
\email{\{youwang.kim, jiyeon.kim, taehyun\}@postech.ac.kr}\\
\url{https://clip-actor.github.io}
}
\date{}

\maketitle

\setcounter{footnote}{0}

This supplementary material aims to provide additional 
contents and details that are not included in the main paper 
due to the space limitation.
In \Sref{sec:supp_dnsf}, we describe the design choices for the 
proposed Decoupled Neural Style Field (DNSF) 
and justify our full model by comparing it with different settings. 
In \Sref{sec:supp_retrieval}, we describe the datasets and experimental details of human motion recommendation.
%
In \Sref{sec:supp_qual}, we provide additional qualitative results of CLIP-Actor.
In \Sref{sec:supp_algorithm}, we provide the overall algorithm for CLIP-Actor.
Furthermore, we provide training details of CLIP-Actor in \Sref{supp:train_details}
and provide the discussion and possible future direction of CLIP-Actor in \Sref{sec:supp_limit}.
We also provide the video results demonstrating the text-conforming stylized meshes in motion.

\section{Analysis on the Decoupled Neural Style Field}
In this section, we introduce our design choices and implementation details 
when composing and optimizing our decoupled neural style field, DNSF, and corresponding effects.
Moreover, we provide implementation details of our mask-weighted embedding attention.

\label{sec:supp_dnsf}
\subsection{Effects of Content Mesh Resolution}
Recall that the CLIP-Actor learns the best text-conforming 
per-vertex color and displacement for the content meshes. 
Thus, the resolution of the content meshes, \ie, the number of mesh vertices, would be the critical factor of the texture generation quality.
Text2Mesh also shows that the na\"ive neural style field network can synthesize a 
more plausible style with high-resolution 
meshes, \ie, meshes with more vertices~\cite{text2mesh}.
%

\begin{figure*}[t]
    \centering
        \includegraphics[width=1.0\linewidth]{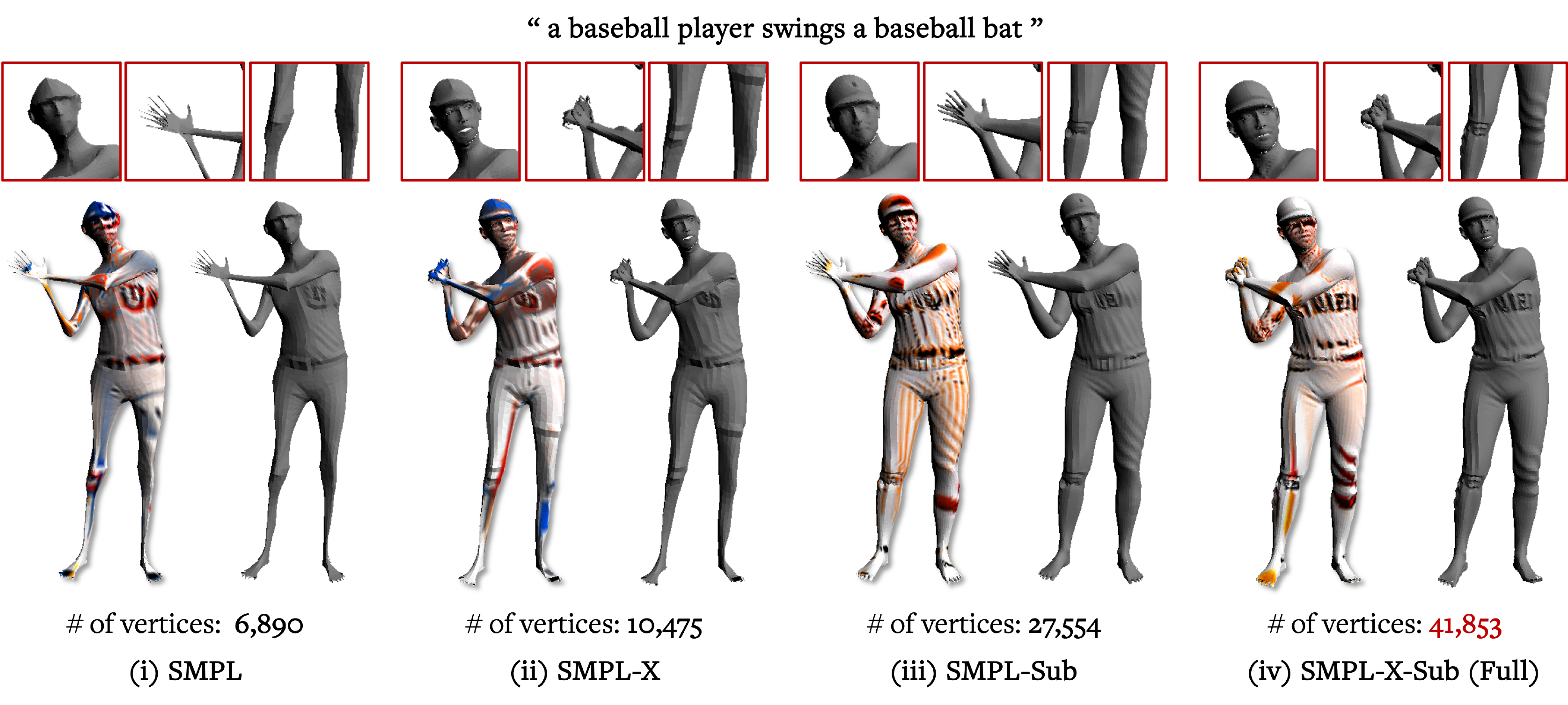}
        \caption{{\textbf{Effects of content mesh resolution.} 
        \textbf{(i)} With SMPL, which has the smallest number of vertices, CLIP-Actor shows unrealistic texture and geometric details. \textbf{(ii)} With higher resolution mesh, SMPL-X, CLIP-Actor achieves much smoother geometry, along with expressive hand and facial details. However, it still suffers from unrealistic colors.
        \textbf{(iii)} With subdivided SMPL, CLIP-Actor achieves better texture and geometry details than (i) and (ii). 
        \textbf{(iv)} CLIP-Actor with subdivided SMPL-X achieves the most realistic color configuration
        and fine-grained geometric details.
        }}
\label{fig:supp_resol}
\end{figure*}

%
\para{SMPL vs. SMPL-X: content mesh selection}
We can change the content mesh model with SMPL~\cite{SMPL:2015} variants, SMPL-H~\cite{MANO:SIGGRAPHASIA:2017} and SMPL-X~\cite{SMPL-X:2019}, which have different numbers of vertices, to investigate the effects of the mesh resolution. 
With its linear blend skinning operation, 
SMPL maps pose parameters, $\mathbf{R}_t$ and shape parameters $\boldsymbol{\beta}_t$, to 6,890 
mesh vertices, \ie, $\mathbf{M}_{\text{SMPL}}\in\mathbb{R}^{6890\times3}$. 
SMPL-X, on the other hand, has 10,475 vertices \ie, $\mathbf{M}_{\text{SMPLX}}\in\mathbb{R}^{10475\times3}$. 
Furthermore, SMPL-X can express detailed hand poses and expressive faces, which is essential in 
modeling human interactions and expressions~\cite{SMPL-X:2019}.
We conduct experiments that compare the qualitative mesh stylization results 
with both mesh models as the content mesh.
Figure~\ref{fig:supp_resol}\colorref{(i),(ii)} illustrate qualitative results of CLIP-Actor with different body mesh models.
With the lowest mesh resolution, \ie, SMPL, CLIP-Actor 
generates
unrealistic body configurations,
such as sharp edges and slim body parts. 
Also, using SMPL as the content mesh, CLIP-Actor cannot represent detailed human actions.
In~\Fref{fig:supp_resol}\colorref{(i)}, the SMPL mesh spreads its hands, thus fails to express the baseball player grabbing the bat.
On the other hand, SMPL-X, which has a higher resolution than SMPL, 
shows a smoother result, detailed hand pose, and facial expressions. 

\para{Mesh subdivision for higher resolution}
Furthermore, we utilize mesh subdivision~\cite{ravi2020pytorch3d}
to achieve higher mesh resolution ($\sim 4\times$ number of vertices). 
Note that we use subdivided SMPL-X for the content mesh for our full model\footnote{Note that we denote our content mesh as SMPL in the main paper for simplicity.}.

Using subdivided meshes of SMPL and SMPL-X (\Fref{fig:supp_resol}\colorref{(iii),(iv)}) as the content mesh,
they show detailed cloth geometry, texture generation and smooth body curvatures than the basic SMPL, SMPL-X meshes. 
Also, SMPL-X-Sub, which is our CLIP-Actor's full version, shows the most realistic color configuration
compared to other mesh models. 
Since our text-driven optimization of decoupled neural style field is based on the rendering of the stylized 
meshes, we postulate that higher resolution of the content meshes results in better supervision signal,
thus leading to improved qualitative results.

%

%



\subsection{Mask-weighted Embedding Attention} 
In the main paper, we mentioned that 2D augmentations are essential for 
plausible texture generation.
Recall that, we apply differentiable 2D augmentations before the rendered images are passed into the pre-trained CLIP encoder.

In practice, we adopt the multi-level 2D augmentations for the 
rendered images, following Text2Mesh~\cite{text2mesh}.
The multi-level 2D augmentation is the method that renders both colored mesh and de-colorized mesh into
images $\mathbf{I}^{*}$ and $\mathbf{I}^{*}_{geo}$, 
computes semantic loss for each rendered image, and leverage gradient accumulation during optimization. 
%
%
%
The advantages of such multi-level 2D augmentations are in two-folds.
First, rendered images in diverse viewpoints and augmentations improve generalization across views~\cite{jain2021dreamfields}. 
Next, separate rendering of textured and de-colorized meshes,
$\mathbf{I}^{*}$ and $\mathbf{I}^{*}_{geo}$, and gradient accumulation
enable guiding both global context and local geometric details with only a single text prompt~\cite{text2mesh}.
%
%

In detail, we apply a global 2D augmentation $\mathcal{T}_{global}(\cdot)$ to the rendered images $\mathbf{I}^{*}$. 
$\mathcal{T}_{global}(\cdot)$ does not contain the image crop but only random perspective transformation.
Also, the local 2D augmentation $\mathcal{T}_{local}(\cdot)$ is applied 
to $\mathbf{I}^{*}$ and $\mathbf{I}^{*}_{geo}$. 
$\mathcal{T}_{local}(\cdot)$ contains both random crops up to 10\% of the original image
and the random perspective transformation.

However, the problem occurs in careless $\mathcal{T}_{local}(\cdot)$.
Prior work~\cite{text2mesh} simply applied
extreme close-ups to the de-colorized rendering of the meshes,
which leads to random, empty rendered images.
Such empty images do not conform to the text prompt, 
and these dummy images can distract the optimization process with random gradient direction.
In CLIP-Actor, we mitigate this problem with mask-weighted attention embedding.

\section{Datasets of Human Motion Recommendation}
\label{sec:supp_retrieval}
In this section, we explain the dataset used in the retrieval system and evaluation for human motion recommendation. 
Moreover, we provide the details of the dataset and the experiment settings.

\para{Retrieval dataset}
We use BABEL~\cite{BABEL:CVPR:2021} as a database of the retrieval system.
BABEL is a dataset that labels a large-scale human motion capture dataset~\cite{AMASS:ICCV:2019} with unique action categories.
Although they provide over 250 action categories, \eg, arm movements, the categories are too abstracted to be matched with our natural language prompt.
Therefore, we utilize the raw labels untrimmed and diverse, \eg, walk without energy and walk fast, 
so that the variants of natural language text prompt can be semantically matched with the raw labels.
%
Instead of using a limited number of closed-set action categories, 
the raw labels can handle the open-set action descriptions.

\para{Evaluation dataset}
To evaluate the text-driven
motion retrieval module, we use the SICK~\cite{Marelli2014ASC} as an evaluation dataset. 
SICK
consists of the sentence pairs obtained from the Flickr8K dataset~\cite{Hodosh2013FramingID} and the video description dataset~\cite{Agirre2012SemEval2012T6}.
Since the sentences in SICK are composed of descriptions of images or video, the dataset is well-matched with our multi-modal retrieval scenario in terms of finding visual semantics. 
%
Each sentence pair in SICK is annotated with a relatedness score from 1 to 5 that indicates the degree of semantic relatedness between two sentences.
We set a range of scores from 4.4 to 4.8 for evaluation settings to ignore unreliable pairs and exclude the pairs that are only different with grammatical voice or article.
SICK4.8 and SICK4.4 settings are constructed with the sentences with the score 4.8 and 4.4, respectively and SICK[4.4,4.8] setting comprises the sentences in the range. 
The samples of the SICK according to the score are shown in Table \colorref{1b} in the main paper.
%
%

\section{Additional Qualitative Results}
\label{sec:supp_qual}
\begin{figure*}[h!]
    \centering
        \includegraphics[width=0.95\linewidth]{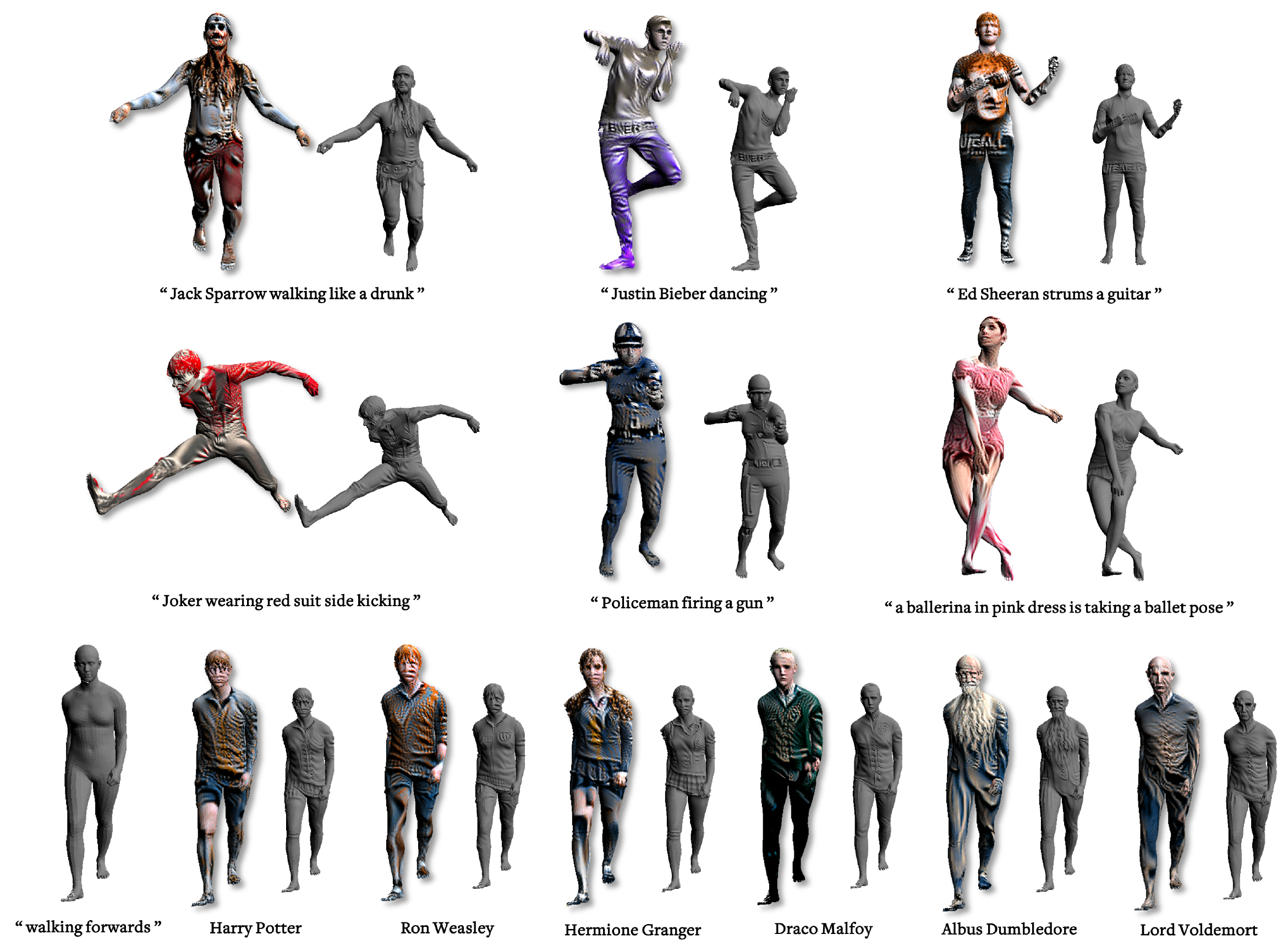}
        \caption{{\textbf{Additional qualitative results.} 
        The first two rows show that CLIP-Actor recommends the human motion from the text description and generates plausible mesh stylization in zero-shot. 
        The last row shows the compositional mesh stylization that allows users to stylize the same retrieved motion, \emph{``walking forwards"} with different identities via text prompt, \eg, Hermione Granger.
        }}
\label{fig:supp_qual}
\end{figure*}

In this section, we present additional qualitative results of CLIP-Actor, with 
diverse subjects and actions (See \Fref{fig:supp_qual})
We describe the text prompts we used and the corresponding results.
Since we cannot express dynamic action sequences in images, 
video results are also attached in the supplementary files.

\clearpage
\section{CLIP-Actor: Algorithm}
\label{sec:supp_algorithm}
In this section, we provide a thorough algorithm for CLIP-Actor. 
CLIP-Actor is a system that includes the recommendation module, 
text-driven DNSF optimization, and stylization, and we arrange the overall pipeline with the algorithm.\\

\begin{algorithm}[H]
    \label{alg:optim}
    \caption{Overall pipeline of CLIP-Actor}
    \noindent{\small \textbf{Require:} Pre-trained CLIP image encoder $\mathbf{g}(\cdot)$, text encoder $\mathbf{h}(\cdot)$,
    
    \hspace*{\algorithmicindent} Pre-trained MPNet text encoder, $\mathbf{m}(\cdot)$, SMPL Linear Blend Skinning $\mathcal{M}(\cdot)$,
    
    \hspace*{\algorithmicindent} BABEL dataset $\mathcal{A}$, SMPL template mesh $\mathbf{M}_{c}$}

     \begin{algorithmic}[1]
    \Require {\small Natural language text prompt $y$}
    \Ensure {\small Text-conforming stylized meshes in motion $\mathbf{M}^{*}_{1:T}$}

    \vspace{2mm}

    \noindent \grc{\small \# {Text-driven Human Motion Recommendation}}\vspace{1mm}
    \State $\mathcal{S}(\mathbf{x}, \mathbf{y}) \stackrel{\text{def}}{=} \tfrac
    {\mathbf{x}^{\top} \mathbf{y}}
    {
    \lVert \mathbf{x}\rVert_{\scriptscriptstyle 2}
    \lVert \mathbf{y}\rVert_{\scriptscriptstyle 2}
    }$ 
    {\small \Comment{Cosine similarity between two vectors, $\mathbf{x}$, $\mathbf{y}$}}
    
    \vspace{1mm}
    \noindent \grc{\small \# {${\texttt{top-k}[\cdot]}$ returns ${\texttt{top-k}}$ items and indices in tuple}}\vspace{1mm}
    
    \State $[\,\mathcal{A}_{\textrm{k}}, \underline{\hspace{3mm}}\,] \gets \verb|top-k|[\mathcal{S}(\mathbf{h}(a_i), \mathbf{h}(y))],\,\,\, \forall a_i \in \mathcal{A}$
    {\small \Comment{Cross-modal aware matching}}
    \State $ a_{*} \gets \argmax_{a_j\in \mathcal{A}_\textrm{k}} \, \mathcal{S}(\mathbf{m}(a_j), \mathbf{m}(y))$;
    {\small  \Comment{Textual semantic matching}}
    
    \vspace{1mm}
    \noindent \grc{\small \# {Get pose parameters from BABEL dataset with retrieved action label, ${a_{*}}$}}\vspace{1mm}
    \State $\mathbf{R}_{1:T} = [\mathbf{R}_1, \dots, \mathbf{R}_T] \gets \verb|BABEL|(a_{*})$
    \State $\mathbf{M}_{1:T} = \mathcal{M}(\mathbf{R}_{1:T}, \boldsymbol{\beta})$;
    {\small \Comment{Motion sequence of the content meshes}}
    \State $\mathbf{I}_{1:T} \gets \verb|render|(\mathbf{M}_{1:T})$;
    \State $[\,\underline{\hspace{3mm}}, \text{idx}\,] \gets \verb|top-k|[\mathcal{S}(\mathbf{g}(\mathbf{I}_{1:T}), \mathbf{h}(y))]$;
    {\small \Comment{Multi-modal content mesh sampling}}
    
    \vspace{2mm}
    \noindent \grc{\small \# DNSF optimization for $L$ iterations}
    \For{$\text{iter}=1,2,\dots,L$}
    \State $\mathcal{L}_s \gets 0$
        
    \State $\mathbf{c}, \mathbf{d} \gets G_{\theta}(\mathbf{M}_{c})$;
    \Comment{\small Decoupled Neural Style Field}
    
    \For {$i \in \text{idx}$}
        {\small \Comment{Temporal view augmentation}}
        \State $\mathbf{M}^{*}_{i} \gets \verb|texturize|(\mathbf{c},\mathbf{d})$;
        
        \State Sample N camera poses, $\mathbf{p}=[\mathbf{p}_1,\dots,\mathbf{p}_{N}]$
        {\small \Comment{3D Spatial augmentation}}
        \For {$j=1,2,\dots,N$}
            \State ${\mathbf{I}}^{*}_{ij}\gets \verb|render|(\mathbf{M}^{*}_{i}, \mathbf{p}_j)$;
            
            \State $\mathbf{I}^{*}_{ij} \gets \verb|2D_augmentations|({\mathbf{I}}^{*}_{ij})$;
            {\small \Comment{2D Spatial augmentation}}
            \State $w_{ij} \gets \verb|mask_weighted_att|(\mathbf{I}^{*}_{ij})$; 
            
        \EndFor
        
        \State $\mathbf{\bar{g}}(\mathbf{I}^{*}_{i}) = 
         \tfrac{\sum_{j=1}^{N}w_{ij}\mathbf{g}(\mathbf{I}^{*}_{ij})}
    {\sum_{j=1}^{N}w_{ij}}$;
        {\small \Comment{Mask-weighted embedding attention}}
        \State $\mathcal{L}_s \gets \mathcal{L}_s +
        (1 - \mathcal{S}(\mathbf{\bar{g}}(\mathbf{I}_{i}^{*}), \mathbf{h}(y)))
        $;
        
    \EndFor
    
    \State $\theta^{*}\gets$ Update DNSF $G_{\theta}$ parameters, $\theta$
    \EndFor
    
    \vspace{1mm}
    \noindent \grc{\small \# {Test time: Stylization of human meshes in motion}}
    \vspace{1mm}
    \State $\mathbf{c}^{*}, \mathbf{d}^{*} \gets G_{\theta^{*}}(\mathbf{M}_{c})$ 
    \Comment{\small Generate color and geometry with learned DNSF}
    \For {$k=1,2,\dots,T$}
        \State $\mathbf{M}_{1:T}^{*} \gets \verb|texturize|(\mathbf{c}^{*}, \mathbf{d}^{*})$
        \Comment{\small Stylize meshes in motion with $\mathbf{c}^{*}, \mathbf{d}^{*}$}
    \EndFor
    \end{algorithmic}
    
\end{algorithm}   

\clearpage

\section{Training Details}
\label{supp:train_details}
We provide training details of CLIP-Actor, including optimizer, training hardware specifications, and training time.
We use the Adam optimizer with the initial learning rate set to 0.0005 and the learning rate decay factor as 0.9 every 100 iterations. 
We train CLIP-Actor for 1500 iterations using a single NVIDIA TITAN RTX GPU.
Total training takes about 30 minutes to one hour depending on the motion sequence length, and training options such as the number of frames we use for spatio-temporal view augmentation.


\section{Discussion}
\begin{wrapfigure}[12]{r}{0.25\linewidth}
\centering
    \includegraphics[width=\linewidth]{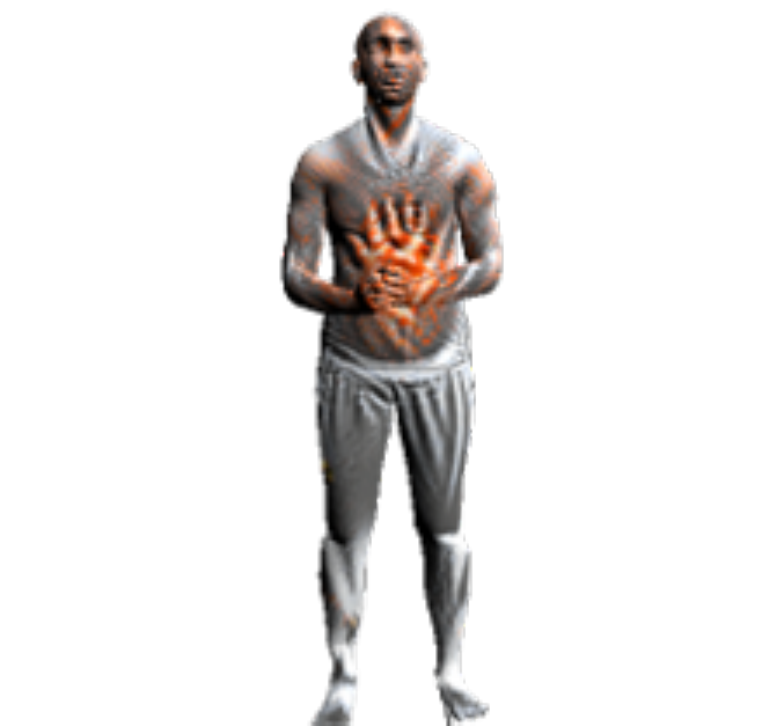}
    \caption{A case of object projection on mesh surface.}
\label{fig:supp_kobe}
\end{wrapfigure}
\label{sec:supp_limit}
We find the observations about human mesh stylization harnessing CLIP~\cite{radford2021learning} text-image joint space.
Given the text prompt that describes an interaction with objects, the objects are often projected onto the human mesh and stylized together.
For example, a basketball is depicted on the player's chest when the action prompt that interacts with the ball is given, \ie, chest passing (see \Fref{fig:supp_kobe}).
Since CLIP is trained with the pairs of text and 2D images, a depth ambiguity from the 2D images can be propagated to the 3D mesh stylization. 
%
The further development of object mesh manipulation can be applied to 
our work to model Human-Object-Interaction~\cite{zhang2020phosa, rhoi2020} as future work.

Since CLIP-Actor recommends the motion conforming to the input prompt instead of generating motions, some prompts might not be compatible with the BABEL~\cite{BABEL:CVPR:2021}.
CLIP-Actor has two major features to prevent such cases.
First, our retrieval module implements semantic matching; 
thus, it finds visual and textual proximal action labels robustly.
For example, given ``Thor swinging Mj\"{o}lnir'' as an input, where ``swinging Mj\"{o}lnir" is 
not included in BABEL, CLIP-Actor retrieves ``swing hammer side to side.'' 

Still, incompatible prompts might exist and harm the subsequent stylization process. 
Our multi-modal content mesh sampling
handles such cases. 
It finds the best mesh frames within the motion to achieve reasonable stylization quality.
%
Our design choices on modules prevent the drastic degradation in stylization even with incompatible prompts.
We think that further improvements to handle out-of-distribution cases would be an interesting future direction.

\clearpage

\end{document}